\pdfoutput=1

\documentclass[11pt]{article}

\usepackage[preprint]{acl}
\usepackage{xurl}
\usepackage{float}
\usepackage{times}
\usepackage{latexsym}
\usepackage{yhmath}
\usepackage{tablefootnote}
\usepackage{amssymb}
\usepackage[T1]{fontenc}

\usepackage[utf8]{inputenc}

\usepackage{microtype}

\usepackage{inconsolata}

\usepackage{graphicx}
\usepackage{float}

\usepackage{microtype}
\usepackage{enumitem}
\usepackage{inconsolata}
\usepackage{xspace}
\usepackage{graphicx}
\usepackage{amsthm}
\newtheorem{assumption}{Assumption A\hspace{-3pt}}
\newtheorem{definition}{Definition D\hspace{-3pt}}
\newtheorem{condition}{Requirement R\hspace{-3pt}}

\usepackage{latexsym}
\usepackage{amsmath} 
\usepackage{amsfonts}
\usepackage{multirow}

\newlist{steps}{enumerate}{1}
\setlist[steps, 1]{label = Source \arabic*:}
\title{Does Data Contamination Detection Work (Well) for LLMs?\\ A Survey and Evaluation on Detection Assumptions}

\author{
 \textbf{Yujuan Velvin Fu\textsuperscript{1}},
 \textbf{\"Ozlem Uzuner\textsuperscript{2}},
 \textbf{Meliha Yeti\c{s}gen\textsuperscript{1}},
 \textbf{Fei Xia\textsuperscript{1}}\\
 \textsuperscript{1}University of Washington,
 \textsuperscript{2}George Mason University
\\
\texttt{\{velvinfu, melihay, fxia\}@uw.edu} \\
\texttt{ouzuner@gmu.edu}
}

\begin{document}
\maketitle 
\begin{abstract}
Large language models (LLMs) have demonstrated great performance across various benchmarks, showing potential as general-purpose task solvers. However, as LLMs are typically trained on vast amounts of data, a significant concern in their evaluation is data contamination, where overlap between training data and evaluation datasets inflates performance assessments. Multiple approaches have been developed to identify data contamination. These approaches rely on specific assumptions that may not hold universally across different settings. To bridge this gap, we systematically review 50 papers on data contamination detection, categorize the underlying assumptions, and assess whether they have been rigorously validated.  We identify and analyze eight categories of assumptions and test three of them as case studies. Our case studies focus on detecting direct, instance-level data contamination, which is also referred to as Membership Inference Attacks (MIA). Our analysis reveals that MIA approaches based on these three assumptions can have similar performance to random guessing, on datasets used in LLM pretraining, suggesting that current LLMs might learn data distributions rather than memorizing individual instances. Meanwhile, MIA can easily fail when there are data distribution shifts between the seen and unseen instances \footnote{Links to all relevant papers and the code for the case study are available on our project GitHub: 
\url{https://github.com/velvinnn/LLM_MIA}.}. 
%


\end{abstract}

\section{Introduction}

Large language models (LLMs) have achieved remarkable performance across various benchmarks, signaling their potential to revolutionize numerous technical domains as general-purpose problem solvers \cite{achiam2023gpt, llama3}. However, a key concern in accurately evaluating those LLMs is the possibility of \textbf{data contamination}, where the LLM's training data overlaps with the evaluation dataset \cite{balloccu2024leak}. Evaluating LLMs on contaminated benchmarks leads to inflated performance assessments \cite{balloccu2024leak,sainz2023nlp,li2024task}, and creates a misleading perception of their capabilities.
Therefore, multiple detection approaches have been developed to identify data contamination in LLMs, and these approaches can also be deployed to identify the use of copyrighted or sensitive content in LLM training \cite{xu2024benchmark, meeuscopyright}.

All existing approaches for detecting data contamination in language models (LMs) rely on specific assumptions regarding the LMs and datasets, which may not be universally applicable across different settings\footnote{This work surveys data contamination detection approaches for LMs of all sizes, not limited to LLMs.}. While previous surveys have focused on detection and mitigation techniques, to our best knowledge, there is currently no comprehensive analysis that surveys and validates the assumptions underlying these approaches \cite{xu2024benchmark, ishihara2023training, hu2022membership}. 

To bridge this gap, we (1) systematically review 50 papers on data contamination detection for LMs, (2) present the formal, mathematical definitions for different levels of data contamination, 
(3) categorize the underlying requirements and assumptions associated with each approach and critically assess whether these assumptions have been rigorously validated, and (4) demonstrate through case studies that some unverified assumptions can be wrong in multiple scenarios. 

\section{Literature Evaluation}\label{sec:paper_collection}
To systematically investigate approaches for data contamination detection, we implement a three-step literature review process with 3 sources of papers: 
(1) four key survey papers \cite{xu2024benchmark, ishihara2023training, hu2022membership, deng-etal-2024-unveiling} and papers from the 1st Workshop on Data Contamination at ACL 2024\footnote{\url{https://conda-workshop.GitHub.io/}.}; 
(2) relevant papers cited by the papers from Source (1); 
and (3) relevant papers cited by the papers from Source (2).

From the above three sources, we collect 81 relevant papers on data contamination. We additionally filter these papers according to the inclusion criterion: the paper must propose and/or evaluate detection approaches for data contamination in text datasets and LMs.
Further more, eight studies solely discussing risks and mitigation strategies for data contamination do not meet the inclusion criteria and are excluded. 

Consequently, our review includes a total of 50 papers. Among them, 
we systematically summarize those detection approaches and present formal mathematical representations for their underlying requirements and assumptions. We then evaluate whether these underlying assumptions are true under different scenarios, as 
described below. 


\section{Levels of Data Contamination}

Data contamination can occur at instance or dataset levels and the detection approaches for them can be different. To facilitate the discussion, we would like to first provide formal mathematical definitions of data contamination at these levels, based on the descriptive definitions from previous research \cite{xu2024benchmark,balloccu2024leak,ishihara2023training}.

\subsection{Instance-Level Contamination}

In this study, we focus on text datasets and define a language instance \( x \) as a sequence of word tokens. Originally, \textbf{direct} instance-level contamination is defined as the presence of an instance $x$ within an LM $M$'s training set, $D_M$, i.e. \( x \in D_M \) \cite{xu2024benchmark}. However, LMs often do not publish their exact training corpus, but instead refer to multiple datasets, as subsets of $D_{M}$ \cite{zhao2023survey}. These datasets typically undergo various pre-processing steps, such as de-duplication, filtering, masking, and removing noise. Consequently, LMs are trained on slightly different versions of the same dataset \cite{palavalli-etal-2024-taxonomy}. Meanwhile, there is also \textbf{indirect} instance-level contamination from greater variations of the dataset, such as machine paraphrasing \cite{yang2023rethinking}. 

To account for such minor differences and indirect contamination, in our Definition D\ref{def:instance_contamination} below, we introduce a \textbf{Binary Indicator Function for Instance-Level Contamination}, $b(x,x')$, which returns \textit{True} (1) if two instances are considered to
be the same and \textit{False} (0) otherwise. Researchers can determine what instances are considered to be the same by defining $b(x,x')$ accordingly.


\begin{definition}\label{def:instance_contamination}
\normalfont \textbf{Instance-Level contamination}: 
Let \( D_{M} \) be the training data of an LM \( M \). The binary function 
$f(M,x)$ is defined as follows:
\end{definition}
 \vspace{-15pt}
 
 \begin{equation}\label{eq:instance_contamination}
        f(M,x) = \left\{
    \begin{array}{ll} 1 & \text{if } \exists x'\in D_M, b(x,x') = 1   \\
                   0 & \text{if } \forall x'\in D_M, b(x,x') = 0
    \end{array}
    \right.
 \end{equation}

We define an instance $x$ to be \textbf{seen} by $M$, or $M$ is \textbf{contaminated} by $x$, iff $f(M,x) = 1$. Conversely, we define an instance $x$ as \textbf{clean} or \textbf{unseen} by $M$ iff $f(M,x) = 0$.

The detection of instance-level contamination is commonly referred to as \textbf{membership inference attack (MIA)}. The goal of MIA is to determine the probability of an instance being used to train an LM, namely, $\widehat{f}(M,x)$ \cite{hu2022membership}. 




\begin{table*}[h]
\centering
\resizebox{0.99\textwidth}{!}{

\begin{tabular}{lllll}
\hline
\textbf{\begin{tabular}[c]{@{}l@{}}Detection\\ Approach\end{tabular}} & \textbf{\begin{tabular}[c]{@{}l@{}}Require-\\ ments (ID)\end{tabular}} & \textbf{\begin{tabular}[c]{@{}l@{}}Assump- \\ tions (ID)\end{tabular}} & \textbf{\begin{tabular}[c]{@{}l@{}}Detection Rearch\end{tabular}} & \textbf{\begin{tabular}[c]{@{}l@{}}Critiques on\\ Those Approaches\end{tabular}} \\ \hline
\begin{tabular}[c]{@{}l@{}}Instance\\ Similarity \\ (9 papers) \end{tabular} & \begin{tabular}[c]{@{}l@{}}\underline{\textcolor{orange}{Disclose}} \\ (R\ref{condi:disclose}) \\ \& \underline{\textcolor{orange}{Release}}\\ (R\ref{condi:accessible})\end{tabular} & None & \begin{tabular}[c]{@{}l@{}}\citet{dodge2021documenting}, \citet{elangovan2021memorization}, \\ \citet{li2023open}, \citet{riddell2024quantifying}, \\ \citet{deng2024investigating}, \citet{yang2023rethinking}, \\ \citet{piktus2023roots}, \citet{lee2022deduplicating},\\ \citet{NEURIPS2023_3112ee70}\end{tabular} &  \\ \hline

\multirow{8}{*}{\begin{tabular}[c]{@{}l@{}}Prob.\\ Analysis \\ (16 papers)\end{tabular}} &   None & \begin{tabular}[c]{@{}l@{}}\underline{\textcolor{orange}{Absolute}}\\ \underline{\textcolor{orange}{Prob.}} (A\ref{as:prob_absolute})\end{tabular} &  \begin{tabular}[c]{@{}l@{}}\citet{10.1145/3292500.3330885}, \\ \citet{shi2023detecting}, \citet{meeuscopyright},  \\ \citet{maini2024llm}, \citet{wei-etal-2024-proving},  \\ \citet{srivastava2023beyond}, \citet{li2023estimating} \end{tabular} & \begin{tabular}[c]{@{}l@{}}\citet{dekoninck2024evading}, \\ \citet{duan2024membership}, \\ \citet{maini2024llm}, \\ \citet{cao2024concerned}, \\ \citet{meeus2024sok} \end{tabular}  \\ \cline{2-5} 
&\begin{tabular}[c]{@{}l@{}}Perturbed \\ Instance (R\ref{condi:ref_sent})\end{tabular} & \begin{tabular}[c]{@{}l@{}}Ref.\\ Prob.  (A\ref{as:prob_reference_sent})\end{tabular} & \begin{tabular}[c]{@{}l@{}}\citet{mattern2023membership}, \citet{maini2024llm} \\ \citet{oren2024proving} \end{tabular}& \begin{tabular}[c]{@{}l@{}} \citet{duan2024membership}, \\ \citet{maini2024llm}, \\ \citet{meeus2024sok} \end{tabular} \\ \cline{2-5} 
 & \begin{tabular}[c]{@{}l@{}}Ref.\\ LM (R\ref{condi:ref_model})\end{tabular} & \begin{tabular}[c]{@{}l@{}}Ref.\\ Prob. (A\ref{as:prob_normalized_sent})\end{tabular} & \begin{tabular}[c]{@{}l@{}}\citet{carlini2021extracting},  \citet{maini2024llm}, \\ \citet{mireshghallah-etal-2022-quantifying},\\ \citet{zanella2020analyzing} \\ \citet{meeus2024did} \end{tabular} & {\begin{tabular}[c]{@{}l@{}}\citet{dekoninck2024evading},\\ \citet{duan2024membership}, \\ \citet{cao2024concerned}, \\ \citet{maini2024llm}, \\ \citet{meeus2024sok}\end{tabular}} \\ \cline{2-5} 
 & Other & Other & \begin{tabular}[c]{@{}l@{}}\citet{jagannatha2021membership}, \citet{zhang2023counterfactual}\end{tabular} &  \\ \hline

\multirow{10}{*}{\begin{tabular}[c]{@{}l@{}}Instance\\ Gen.\\ \&\\ Instance\\ Select. \\
(20 papers)\end{tabular}} & None & \begin{tabular}[c]{@{}l@{}}\underline{\textcolor{orange}{Verbatim}}\\ \underline{\textcolor{orange}{Mem.}}  (A\ref{as:exact_memorization})\end{tabular} & \begin{tabular}[c]{@{}l@{}}\citet{carliniquantifying}, \citet{kandpal2022deduplicating},\\ \citet{magar2022data}, \citet{duarte2024decop}*, \\ \citet{tirumala2022memorization}, \citet{schwarzschild2024rethinking}, \\\citet{Golchin2023quiz}*\end{tabular} &  \\ \cline{2-5} 
 & \begin{tabular}[c]{@{}l@{}}Key\\ Info. \\ (R\ref{condi:key_info})\end{tabular} & \begin{tabular}[c]{@{}l@{}}Key\\ Info.\\ Gen. (A\ref{as:key_memorization})\end{tabular} &  \begin{tabular}[c]{@{}l@{}}\citet{deng2024investigating}, \citet{ranaldi2024investigating},\\ \citet{chang2023speak}, \citet{9152761},\\ \citet{carlini2021extracting}, \citet{236216},\\  \citet{liu-etal-2024-evaluating}, \citet{golchintime},\\\citet{Golchin2023quiz}\end{tabular} & 
 \\ \cline{2-5} 
 & None & \begin{tabular}[c]{@{}l@{}}\underline{\textcolor{orange}{Gen.}}\\ \underline{\textcolor{orange}{Variation}} (A\ref{as:Generation_Variation})\end{tabular} & \citet{dong2024generalization} &  \\ \cline{2-5} 
 &  \begin{tabular}[c]{@{}l@{}}Metadata\\ (R\ref{condi:meta_data})\end{tabular} & \begin{tabular}[c]{@{}l@{}}Metadata\\ Mem. (A\ref{as:meta_data})\end{tabular} & \begin{tabular}[c]{@{}l@{}}\citet{sainz2023chatgpt}*\\ \citet{karamolegkou-etal-2023-copyright}*\end{tabular} & \citet{dekoninck2024evading}   \\ \hline
\begin{tabular}[c]{@{}l@{}}Answer\\ Mem. \\ (5 papers)\end{tabular} & \begin{tabular}[c]{@{}l@{}}Instance\\ Perturb. (R\ref{condi:pertub_instance})\end{tabular} & \begin{tabular}[c]{@{}l@{}}Answer\\ Change (A\ref{as:same_task})\end{tabular} &\begin{tabular}[c]{@{}l@{}} \citet{liu-etal-2024-evaluating}*, \citet{mehrbakhsh-etal-2024-confounders}*, \\ \citet{yim-etal-2024-err-human}*, \citet{zong2023fool}*, \\ \citet{razeghi2022impact}*  \end{tabular} & \\  \hline
\end{tabular}
}

\caption{Existing detection approaches for direct data contamination, their requirements and assumptions, and critiques they received. Some papers cover multiple detection approaches with different assumptions. 
Most detection methods apply to both instance- and dataset-level contamination, while * denotes those limited to dataset-level contamination.
In this study, we show that the \underline{\textcolor{orange}{underlined}} assumptions may not be often satisfied.}

\label{tab:summmary}
\end{table*}



\subsection{Dataset-Level Contamination}
Prior research implicitly refers to dataset-level contamination at two degrees: partial dataset contamination and full dataset contamination.

\begin{definition}
    \normalfont \textbf{Full Dataset Contamination}: A dataset $D$ is fully contaminated (\textbf{fully seen}) by an LM, if every instance within this dataset is contaminated. Namely, $\forall x \in D, f(M,x)=1$.
\end{definition}

When creating benchmarks for detecting data contamination, previous work typically generates the fully contaminated split. 
For example, \citet{maini2024llm} created contaminated and clean datasets, respectively from the training and validation splits of the LM's pretraining corpus. \citet{shi2023detecting} focused on LMs which disclosed that they used Wikipedia event data for training, and created the contaminated dataset from the Wikipedia event data which were published before the LMs' release. 

\begin{definition}
\normalfont
\textbf{Partial Dataset Contamination}: A dataset $D$ is partially contaminated  (\textbf{paritially seen}) by an LM $M$, if at least one instance within $D$ is seen. Namely, $\exists x \in D, f(M,x)=1$.
\end{definition}

In practice, especially when reporting contamination from benchmark datasets \cite{dong2024generalization} or detecting copyrighted content \cite{karamolegkou-etal-2023-copyright,chang2023speak}, people focus more on evaluating partial dataset contamination.

\begin{definition}
\normalfont
\textbf{Unseen/Clean Dataset}: A dataset is clean (\textbf{unseen}) by an LM, if none of its instances is contaminated. Namely, $\forall x \in D, f(M,x)=0$.
\end{definition}

\section{Detection of Direct Data Contamination} \label{sec:direct}
Direct data contamination is the most common and well-researched type of data contamination. In this section, we categorize the existing detection approaches, their requirements, assumptions, and the critiques they received (see Table \ref{tab:summmary}). The requirements are defined as the preliminary conditions necessary for conducting certain detection approaches.
The assumptions
are what the authors of detection approaches assume to be true; 
the assumptions either are explicitly stated by the authors or can be inferred from the detection approaches.

Most detection methodologies for direct contamination are primarily developed to address instance-level contamination; however, they can be adapted to account for dataset-level contamination. Consequently, unless specified otherwise, this section will concentrate on instance-level contamination.

The performance of a detection method depends on how well its requirements are met and the reliability of its assumptions. Therefore, we group the detection approaches based on their assumptions and requirements.

\subsection{Instance Similarity}\label{sec_instance_similarity}

When $D_{M}$ is known, detection approaches based on \textbf{instance similarity} directly deploy Equation \ref{eq:instance_contamination}, by proposing a similarity function to measure $b(\cdot,\cdot)$ and comparing a new instance with every $x \in D_M$.  

Previous research focuses on developing a better or more efficient similarity function. 
Examples of similarity calculation can be conducted through exact match \cite{dodge2021documenting}, fuzzy match \cite{piktus2023roots,lee2023platypus}, automatic NLG evaluation metrics \cite{elangovan2021memorization,deng2024investigating}, and another LM \cite{yang2023rethinking}. Tools have also been developed to allow efficient search within a large $D_{M}$, such as Data Portraits \cite{NEURIPS2023_3112ee70} and ROOTS Search Tool \cite{piktus2023roots}. 

Although this approach does not rely on underlying assumptions, it has two requirements:
\begin{condition}\label{condi:disclose}
  \normalfont
    $D_{M}$ needs to be disclosed. 
\end{condition}
\begin{condition}\label{condi:accessible}
  \normalfont
    $D_{M}$ must be accessible, which is often hindered by legal, privacy constraints, and expired website links.
\end{condition}

\textbf{Case Study}: To examine how often these two requirements are met, we analyzed the top 10 LMs on the Vellum LLM leaderboard \footnote{\url{https://www.vellum.ai/llm-leaderboard\#model-comparison}. Accessed on Oct 6, 2024.}. We found that \textit{none} of the LMs fulfilled R\ref{condi:disclose}, the most basic  requirement, let alone R\ref{condi:accessible}
(see Appendix \ref{sec:appendix:conditions} for details).

\subsection{Probability Analysis}\label{sec:prob_analysis}
When the training dataset $D_M$ is unavailable, but the LM $M$'s output token probabilities are known, probability analysis has been used to detect potential instance-level contamination. We group those detection approaches by their assumptions, and unless specified otherwise, they have no requirements. 

\subsubsection{Absolute Probability} \label{sec:abs_prob}
Given an instance $x$, probability analysis measures instance-level contamination through $P_M(x)$, the probability of the instance $x$ based on an LM \( M \).
\begin{assumption}\label{as:prob_absolute}
  \normalfont Seen instances will have higher probabilities than unseen ones, and there exists a threshold, $\xi_p $, that separates seen instances from unseen ones: 
\end{assumption}
\vspace{-15pt}
\begin{equation}
    P_M(x) \left\{
    \begin{array}{ll} \geq \xi_p & \text{if } f(M,x) = 1  \\
                   < \xi_p & \text{if } f(M,x) = 0
    \end{array}
    \right.
 \end{equation}

Previous research measures $P_M(x)$ through perplexity \cite{carlini2021extracting, li2023estimating} or approximates it through LM loss \cite{wei-etal-2024-proving}, which can be impacted by the instance domain and simplicity. 
To improve upon this assumption, \citet{shi2023detecting} 
evaluates only the average token probability of top $p\%$ least likely tokens in an instance (\textbf{Min $p$\% Token}), assuming that unseen instances contain more low-probability outliers in Wikipedia events and books. Similarly, \citet{10.1145/3292500.3330885} assesses probabilities of the $k$ most frequent tokens.

Likewise, \citet{srivastava2023beyond}, \citet{wei-etal-2024-proving}, and \citet{meeuscopyright} proposed inserting special strings as watermarks into the training data, using the probability of these watermarks to detect data contamination.
 
However, \citet{maini2024llm} and \citet{duan2024membership} have demonstrated that the perplexity and min top $p$ probabilities are close to random in detecting direct instance-level data contamination across different splits of the Pile dataset. \citet{maini2024llm} suggests that shifts in perplexity and infrequent word probabilities may be attributed to temporal events on platforms like Wikipedia, rather than contamination. Similarly, \citet{cao2024concerned} highlighted that perplexity and token probability approaches are ineffective for code generation tasks.



\subsubsection{Reference Probability by An Instance}

Instead of assuming the probabilities of all the seen instances are higher than the probabilities of all the unseen instances, this approach compares the probabilities of similar instances. 

\begin{condition}\label{condi:ref_sent}
  \normalfont
  There exists an algorithm which, 
    given an instance $x$ and an LM $M$,  can automatically generate a similar unseen instance,  $x'$.
\end{condition}

\begin{assumption}\label{as:prob_reference_sent}
  \normalfont 
  If  $x$ and $x'$ are similar and $M$ has seen $x$ but not $x'$ , the probability of $x$ should 
  be much higher than that of $x'$ based on M:  
\end{assumption}
\vspace{-15pt}
\begin{equation}
    P_M(x) \left\{
    \begin{array}{ll} \gg  P_M(x') & \text{if } f(M,x) = 1  \\
                   \not\gg  P_M(x') & \text{if } f(M,x) = 0
    \end{array}
    \right.
 \end{equation}
 

Utilizing this assumption, \citet{mattern2023membership} construct the similar, reference instance $x'$ by replacing individual words in $x$ with their synonyms. 
However, in practice, the observation that $P_M(x) \ge P_M(x')$ might result from replacement with rare words. This assumption has been proven false by \citet{maini2024llm} on different splits of the Pile dataset \cite{gao2020Pile}. In addition to this synonym-based perturbation, \citet{maini2024llm} demonstrate the ineffectiveness of other perturbation approaches, including white space, characters, random deletion, and case changes. 

Another study, \citet{oren2024proving}, constructs the reference instance by randomly shuffling (exchanging) the order of sentences in the original instance. They make another assumption of the \textit{exchangeability}, positing that all orderings of all the instances in an exchangeable benchmark should be equally likely if uncontaminated. This assumption might not be valid for coding and reasoning tasks.

\subsubsection{Reference Probability by Another LM}

This type of approach compares the probability of an instance based on two LMs.

\begin{condition}\label{condi:ref_model}
  \normalfont
    Given an instance $x$, we can find another LM $M'$ such that $x$ is unseen by $M'$. 
\end{condition}

\begin{assumption}\label{as:prob_normalized_sent}
  \normalfont 
  If $x$ is seen by $M$ but not $M'$, then
   $P_{M}(x)$ should be much higher than $P_{M'}(x)$:  
\end{assumption}
\vspace{-15pt}
\begin{equation}
    P_M(x) \left\{
    \begin{array}{ll} \gg P_{M'}(x) & \text{if } f(M,x) = 1  \\
                   \not\gg  P_{M'}(x) & \text{if } f(M,x) = 0
    \end{array}
    \right.
 \end{equation}

Previous research has utilized term frequency \cite{meeus2024did},  the zlib entropy \cite{carlini2021extracting}, and another LM \cite{carlini2021extracting, mireshghallah-etal-2022-quantifying} as the reference model. However, \citet{maini2024llm} and \citet{duan2024membership} have demonstrated that those reference models 
perform close to random guessing across various domains. 
\citet{cao2024concerned} also show the zlib entropy does not work for code generation tasks. Instead of using reference probabilities at the sentence level, \citet{zanella2020analyzing} deploy a reference model for both individual token probability and its probability rank within the vocabulary.

\subsection{Instance Generation and Instance Selection} \label{sec:instance_generation}
In this section, we investigate underlying requirements and assumptions for detection approaches based on instance generation and instance selection. 

Instance generation detects contamination by treating $x$ as a prefix-suffix pair, $x = (x_p, x_s)$. These approaches evaluate the LM $M$'s generated output, $M(x_p)$, conditioned on $x_p$. If $M(x_p)$ is similar or identical to $x_s$, $x$ will be predicted as seen. Based on this core intuition, instance generation approaches vary in their assumptions regarding input-output pairs and language generation approaches. Unless specified otherwise, those approaches below focus on instance generation.

For instance selection, instead of directly generating answers, the LM is tasked with selecting the most likely $x_s$ from a set of candidate options in a multi-choice format.
However, detection approaches relying on instance selection face a fundamental limitation: even if $x$ is unseen, the LM might still choose the correct $x_s$ by accident. 
Consequently, these approaches are generally not employed to detect instance-level contamination but rather to assess the probability of full dataset contamination.

\subsubsection{Verbatim Memorization}
This type of approach assumes LMs can memorize their training data, to certain extent.

\begin{assumption}
\label{as:exact_memorization} 
  \normalfont Given an prefix-suffix pair $x = (x_p, x_s)$, if $x$ has been seen by an LM $M$,  $x_s$ can be generated (memorized) by $M$ through greedy decoding, when given the input  $x_p$.
\end{assumption}
\vspace{-15pt}
\begin{equation}
    M_g(x_p) \left\{
    \begin{array}{ll}  = x_s & \text{if}  f(M,x)=1    \\
                      \neq x_s & \text{if}  f(M,x)=0 
    \end{array}
    \right.
 \end{equation}
 

\citet{duarte2024decop} and \citet{Golchin2023quiz} define $x_s$ as sentences or passages, and create similar instances to $x_s$ by paraphrasing $x_s$ using another LM. They use instance selection, and assume that the contaminated LM will be more likely to select the verbatim option. 

However, instance-level contamination does not always lead to verbatim memorization. Utilizing instance generation, \citet{kandpal2022deduplicating}, \citet{236216}, \citet{carliniquantifying}, and \citet{tirumala2022memorization} demonstrate that verbatim memorization requires repeated exposures to this instance $x$ during training, and a larger LM and longer input length $x_p$ can result in better memorization.  
\citet{schwarzschild2024rethinking} used the minimum length of $x_p$ needed to generate the desired output $x_s$ to define the degree of memorization.

Similarly,  \citet{kandpal2022deduplicating} and \citet{carlini2021extracting} study a relaxed version of this assumption, where the LM can generate $x_s$ through different sampling strategies in decoding, such as top-\textit{k} or top-\textit{p} (Nucleus) sampling \cite{holtzman2019curious}. They reach a similar conclusion that data contamination does not necessarily lead to memorization.

\subsubsection{Key Information Generation}
This type of approach assumes that, if an LM has seen an instance, it can generate $x$'s key information based on its context.

\begin{condition}\label{condi:key_info}
  \normalfont
    An instance $x$ can be paraphrased into a slot-filling, context-key pair, $x = (x_c, x_k)$. The key $x_k$ is usually a representative sub-span of $x$,  such as dates and names. The rest tokens in $x$ compose the context, $x_c$.
\end{condition}

\begin{assumption}\label{as:key_memorization}
  \normalfont
   If $x$ is seen, $M$ will be able to produce similar output to $x_k$
  when given $x_c$. 
\end{assumption}
\vspace{-15pt}
\begin{equation}
    S(M(x_c),x_k) \left\{
    \begin{array}{ll} \geq  \tau_s & \text{if } f(M,x) = 1  \\
                   < \tau_s & \text{if } f(M,x) = 0
    \end{array}
    \right.
 \end{equation}
Here,  $M(x_c)$ denotes the output of the LM $M$ through a certain decoding method. $S(\cdot,\cdot)$ is a text similarity function, and $\tau_s$ is the contamination threshold. 
One can use the similarity functions  described in Section \ref{sec_instance_similarity}.

Leveraging this assumption, prior studies have masked key information within specific datasets, including input questions in NLP benchmarks \cite{deng2024investigating, golchintime, liu-etal-2024-evaluating}, column names in SQL code generation questions \cite{ranaldi2024investigating}, character names in books \cite{chang2023speak}, and labels in NLI and SST tasks \cite{magar2022data}. 


\subsubsection{Generation Variation}
This type of approach explores how an LM's outputs vary if it has seen an instance during training.

\begin{assumption}\label{as:Generation_Variation}
  \normalfont Suppose an instance $x$ can be 
  represented as a prefix-suffix pair, $x=(x_p,x_s)$. 
  If an LM $M$ has seen $x$, then given $x_p$, $M$
  will generate something
  identical or similar to $x_s$ under 
  different sampling strategies: 
\end{assumption}
\vspace{-15pt}
\begin{equation}
    \text{Var}(\{M_{\cdot}(x_p)\}) \left\{
    \begin{array}{ll} <  \xi_v & \text{if } f(M,x_p) = 1  \\
                   \geq \xi_v & \text{if } f(M,x_p) = 0
    \end{array}
    \right.
 \end{equation}
where $Var(\{M_{\cdot}(x_p)\})$ measures the variations of outputs from $M$ 
under diverse, different sampling strategies when given $x_p$;
$\xi_v$ is a threshold, based on the type of input $x_p$ and sampling strategies.

\citet{dong2024generalization} defines the metric $Var(\cdot)$ as `Contamination Detection via output Distribution' (CDD), and utilizes this assumption to detect memorization in coding and reasoning benchmarks. However, this assumption can lead to false positives for other tasks, such as multiple choices, where the output is more constrained and has less variation. 

\subsubsection{Metadata-based Memorization}
This type of approach 
determines whether an LM has seen a dataset $D$
by using $D$'s metadata.

\begin{condition}\label{condi:meta_data}
  \normalfont
    Given a dataset $D$, we can construct an input prompt $x_m$ including $D$'s metadata $m$, such as dataset name, split, and format.
\end{condition}

\begin{assumption}\label{as:meta_data}
  \normalfont If an LM $M$ has seen a dataset $D$,
  when given $D$'s metadata, $M$ is able to 
  generate instances that are very similar to some  $x \in D$.
  
\end{assumption}
\vspace{-15pt}

 \begin{equation}
    \left\{
    \begin{array}{ll} 
    \exists x \in D, S(M(x_m),x) \ge \tau_m &  \text{if $D$ is seen} \\
    \forall x \in D, S(M(x_m),x) < \tau_m & \text{if $D$ is unseen}
    \end{array}
    \right.
 \end{equation}

Here, $M(x_m)$ is the set of instances that $M$ generates when given $D$'s metadata $m$; 
$S(M(x_m), x)$ represents the highest similarity
between $x$ and any instance $x'$ as a subsequence of $M(x_m)$; 
$\tau_m$ is the contamination threshold for $S(\cdot,\cdot)$.

\citet{sainz2023chatgpt} and \citet{golchintime} utilized this assumption to demonstrate that OpenAI systems memorized many instances from widely used benchmarks. However, this approach can have false negatives if the LM's training phase does not preserve the linkage between  $D$'s metadata and instances \cite{dekoninck2024evading}.

\subsection{Answer Memorization}
Answer memorization is usually conducted at the dataset level. It introduces perturbations to the original dataset, measures the LM's performance change, and aims to detect whether the LM's high performance is due to memorizing its answer.

\begin{condition}\label{condi:pertub_instance}
  \normalfont Given an LM $M$ and an evaluation dataset $D$,
    one can generate a similar dataset $D'$ that is unseen by $M$, by modifying every $x \in D$.
\end{condition}

\begin{assumption}\label{as:same_task}
  \normalfont  Suppose datasets $D$ and $D'$ are
  similar and an LM $M$ has seen $D$ but not $D'$, 
  $M$'s performance on $D$ ($\text{Eval}(M,D)$) will be much higher than
  its performance on $D'$ ($\text{Eval}(M,D')$). 
\end{assumption}

\vspace{-15pt}
                   

\begin{equation}\label{equal:task_contamination}
    \text{Eval}(M,D) \left\{
    \begin{array}{ll} \gg  \text{Eval}(M,D')& \text{if $D$ is seen} \\
    \not\gg  \text{Eval}(M,D') & \text{if $D$ is unseen} 
    \end{array}
    \right.
 \end{equation}

Previous research evaluates answer memorization in multiple-choice (MC) tasks by introducing variations such as altering numbers in mathematical tasks \cite{mehrbakhsh-etal-2024-confounders},  changing the order of MC options, etc. \cite{yim-etal-2024-err-human, zong2023fool}. \citet{razeghi2022impact} show that multiple LMs perform better on numerical reasoning tasks involving frequently occurring numerical variables in their pretraining data. 
Similar to this assumption, \citet{liu-etal-2024-evaluating} detects if the LM can still predict the correct answer, after removing all MC options.

\section{Other Types of Contamination}
Besides direct data contamination, previous research also investigates indirect data contamination (6 papers) and task contamination (5 papers).

\subsection{Indirect Data Contamination}

Indirect data contamination occurs when an instance \( x \) is not seen by an LM $M$, but something (\( x'\)) derived from x  is. For instance, \( x'\) can be a paraphrase or a summary of \( x\) \cite{yang2023rethinking}.

Indirect data contamination is often hard to track and trace \cite{balloccu2024leak}. 
For example, OpenAI uses online user conversations for training, which could include variations of benchmark datasets\footnote{\url{https://help.openai.com/en/articles/5722486-how-your-data-is-used-to-improve-model-performance}. Accessed on Oct 6, 2024.}. Another example involves knowledge distillation, where an LM utilizes instances $x_k$ generated by another LM $M'$ during training, and these instances $x_k$ may resemble instances from the training set $D_{M'}$ of $M'$ \cite{veselovsky2023artificial}.


\subsubsection{Detection Approaches for Indirect Data Contamination}
Compared to direct contamination, indirect data contamination is much more challenging to detect. \citet{dekoninck2024evading} and \citet{cao2024concerned} show that many probability-based detection approaches are ineffective for indirect data contamination. 

However, prior research showed that three approaches may still be applicable: (1) the instance similarity measured by another LM \cite{yang2023rethinking}, (2) the CDD metric \cite{dong2024generalization}, which leverages Assumption A\ref{as:Generation_Variation} by measuring output variations rather than directly comparing with original instances, and (3) directly tracking the disclosed usage of datasets. For example, \citet{balloccu2024leak} reviewed the datasets evaluated using OpenAI APIs. 


\subsection{Task Contamination}  \label{sec:task_contamination}
Task contamination occurs when any instance of the same task is seen by an LM \cite{li2024task}. Detecting task contamination is crucial for assessing an LM's generalizability to unseen tasks \cite{chung2024scaling}. Tasks can include applications such as machine translation, summarization, and mathematical calculation. Task contamination is a broader concept than data contamination: if some labeled instances from a dataset are seen by an LM, the associated task is contaminated, but task contamination doesn't always imply the dataset has been seen.

Task contamination generally evaluates an LM's performance on a particular task at the dataset level. The idea is that if an LM has previously seen the task, its performance will be much higher compared to unseen tasks of similar difficulty.

For example, as noted by \citet{aiyappa-etal-2023-trust}, LLMs show improved performance on the same benchmark after model updates, which may be influenced by data contamination during LLMs' continuous training. 
\citet{ranaldi2024investigating} and \citet{li2024task} also find that OpenAI models perform significantly better on benchmarks released before the model's release than on those released later, when task difficulty is controlled or performance is normalized using a baseline model. To ensure fair comparisons across tasks, \citet{li2024task} control task difficulty using a baseline system. However, \citet{cao2024concerned} also note that LMs do not necessarily perform worse on more recent code generation benchmarks.


\section{Case Study}\label{sec:case_study}
Besides the case study in Section \ref{sec_instance_similarity}, we aim to evaluate whether the assumptions outlined in Table \ref{tab:summmary} are universally applicable across different domains, for direct and instance-level MIA.

\begin{table*}
\centering
\resizebox{1\textwidth}{!}{
    \begin{tabular}{lllllll}
    \hline
    \textbf{LM} & \textbf{\begin{tabular}[c]{@{}l@{}}\#\\ Params\end{tabular}} &
    \textbf{\begin{tabular}[c]{@{}l@{}}Training\\ Phase\end{tabular}}&
    \textbf{\begin{tabular}[c]{@{}l@{}}\#\\ Epochs\end{tabular}} &
    \textbf{\begin{tabular}[c]{@{}l@{}}Trainset\\ Size\end{tabular}} &
    \textbf{\begin{tabular}[c]{@{}l@{}}Batch\\ Size\end{tabular}} &
    \textbf{\begin{tabular}[c]{@{}l@{}}Seen \& Unseen Datasets \\ Used in Our Case Study\end{tabular}} \\ \hline

    Pythia \cite{biderman2023pythia} 
    & 70M - 12B  &\multirow{2}{*}{Pretraining} 
    & $\approx$  1.5 & 825 GiB & 2M & Pile \cite{gao2020Pile} \\\cline{1-2} \cline{4-7} 
    
    OLMo-2  \cite{olmo20242olmo2furious}  & 7B &  & $\approx$ 2 & 22.4 TB & $\approx$ 4M & AlgebraicStack  \cite{azerbayev2023llemma} \\ \hline
    
    Zephyr-7B-$\beta$  \cite{tunstall2023zephyr}  & 7B & \multirow{2}{*}{\begin{tabular}[c]{@{}l@{}} Supervised \\ Fine-tuning\end{tabular}} & 1-3 & 9.3 GB & 512 & UltraChat \cite{ding2023enhancing} \\ \cline{1-2} \cline{4-7} 
    
    BioMistral-NLU \cite{fu2024biomistral} & 7B &  & 2 & 3.6 GB & 64 & Medical-NLU \cite{fu2024biomistral}\\ \hline
    \end{tabular}
}
\vspace{-5pt}
\caption{LMs and datasets used in the case study. Except for the UltraChat dataset, each dataset contains multiple subsets from different domains. The trainset refers to the whole trainset used in each LM's corresponding training phase, as described in their original paper, which is a superset of seen \& unseen datasets used in our case study.}
\label{tab:models_and_datasets}
\vspace{-5pt}
\end{table*}

\begin{table*}[h]
\centering
\resizebox{0.98\textwidth}{!}{

\begin{tabular}{llccccccc}
\hline
\multicolumn{2}{l|}{\textbf{Training Phase}} & \multicolumn{4}{c|}{\textbf{Pretaining}} & \multicolumn{3}{c}{\textbf{Supervised Fine-tuning}} \\ \hline
\multicolumn{2}{l|}{\textbf{Model}} & \multicolumn{2}{c|}{\textbf{Pythia-6.9B}} & \multicolumn{2}{c|}{\textbf{OLMO-2-7B}} & \multicolumn{1}{c|}{\textbf{Zephyr-7B-$\beta$ }} & \multicolumn{2}{c}{\textbf{BioMistral-NLU-7B}} \\ \hline

\multicolumn{2}{c}{\textbf{Assumptions \& Metric}} & \textbf{\begin{tabular}[c]{@{}c@{}}Youtube-\\ Subtitles\end{tabular}} & \textbf{ArXiv} & \textbf{\begin{tabular}[c]{@{}c@{}}Github-\\ Coq\end{tabular}} & \textbf{\begin{tabular}[c]{@{}c@{}}Github-\\ Isabelle\end{tabular}} & \textbf{\begin{tabular}[c]{@{}c@{}}Ultra\\ Chat\end{tabular}} & \textbf{\begin{tabular}[c]{@{}c@{}}DC-\\ MTSample\end{tabular}} & \textbf{\begin{tabular}[c]{@{}c@{}}RE-\\ 2012temp\end{tabular}} \\ \hline
 & PPL\_50 & 50.7 & 50.7 & 47.1 & 50.9 & 55.3 & 51.9 & {\color[HTML]{38761D} \textbf{62.5}} \\
 & PPL\_100 & 50.4 & 51.1 & 48.2 & 53.1 & 59.1 & 60.0 & {\color[HTML]{38761D} \textbf{95.3}} \\
 & PPL\_200 & 49.6 & 50.9 & 48.5 & 51.6 & {\color[HTML]{38761D} \textbf{60.1}} & 58.9 & {\color[HTML]{38761D} \textbf{99.4}} \\ \cline{2-2}
 & Min 5\% token & 48.5 & 51.3 & 49.4 & 54.0 & {\color[HTML]{38761D} \textbf{63.6}} & 47.4 & {\color[HTML]{38761D} \textbf{92.9}} \\
 & Min 15\% token & 48.6 & 51.3 & 49.3 & 53.6 & {\color[HTML]{38761D} \textbf{63.0}} & 51.7 & {\color[HTML]{38761D} \textbf{93.3}} \\
\multirow{-6}{*}{A\ref{as:prob_absolute}} & Min 25\% token & 48.5 & 51.4 & 49.3 & 53.1 & {\color[HTML]{38761D} \textbf{61.5}} & 53.3 & {\color[HTML]{38761D} \textbf{93.4}} \\ \cline{1-2}
 & Mem 5 & 49.1 & 52.7 & 52.1 & 48.7 & 52.2 & 47.9 & 41.4 \\
 & Mem 15 & 48.6 & 51.9 & 50.6 & 58.6 & 53.1 & 49.1 & 45.2 \\
\multirow{-3}{*}{A\ref{as:exact_memorization}} & Mem 25 & 48.2 & 51.6 & 49.4 & 59.2 & 53.1 & 50.3 & 48.9 \\ \cline{1-2}
 & Entropy 5 & 49.3 & 52.0 & 47.2 & 52.3 & 54.3 & 55.4 & {\color[HTML]{38761D} \textbf{93.2}} \\
 & Entropy 15 & 49.0 & 52.0 & 47.8 & 52.1 & 54.1 & 54.8 & {\color[HTML]{38761D} \textbf{93.1}} \\
\multirow{-3}{*}{A\ref{as:Generation_Variation}} & Entropy 25 & 48.9 & 51.9 & 48.6 & 52.5 & 54.0 & 54.1 & {\color[HTML]{38761D} \textbf{93.1}} \\ \hline
\multicolumn{2}{c}{\textbf{Average AUC}} & 49.0 & 51.8 & 49.0 & 53.3 & 55.9 & 52.0 & {\color[HTML]{38761D} \textbf{74.1}} \\ \hline
 & Seen & 13.2$\pm$16.0 & 7.9$\pm$3.7 & 10.5$\pm$8.1 & 8.4$\pm$5.2 & 5.3$\pm$4.6 & 1.7$\pm$0.2 & 1.4$\pm$0.1 \\
\multirow{-2}{*}{PPL} & Unseen & 12.7$\pm$10.6 & 8.0$\pm$3.6 & 9.9$\pm$7.2 & 9.2$\pm$6.5 & 6.2$\pm$4.1 & 1.8$\pm$0.2 & 3.0$\pm$1.6 \\ \hline

\end{tabular}
}
\vspace{-5pt}
\caption{Average MIA AUC for different LMs. For LMs evaluated on multiple subsets (domains) of the same dataset, we present the results from the subsets with the lowest and highest average AUC. The last two rows, marked as `PPL\_200', represent the average perplexity $\pm$ STD, from the first 200 tokens within every instance. The color {\color[HTML]{38761D} green } represents AUCs higher than 60.}
    \label{tab:overall}
    \vspace{-10pt}
    \end{table*}
    
\subsection{Assumptions to Evaluate}
As shown in Table \ref{tab:summmary}, prior research has verified that 4 out of 8 assumptions can fail under certain conditions. Meanwhile, some assumptions have specific requirements, and their applicability depends on how well these requirements are met. Therefore, we focus on two unverified assumptions that have no such requirements for evaluation, limiting confounding factors and deferring the testing of other assumptions to future studies. We also validate one verified assumption (Assumption A\ref{as:prob_absolute}) to confirm the consistency of our findings with prior research.

\textbf{Assumption A\ref{as:prob_absolute}: Absolute Probability}. In the assumption that seen instances will have a lower perplexity and fewer low-likely (outlier) tokens, we measure perplexity by an instance's first $k$ tokens (\textbf{PPL\_$k$}) \cite{carlini2021extracting};  \textbf{Min $p$\% Token} by the average token probabilities among $p\%$ least likely tokens \cite{shi2023detecting}.

\textbf{Assumption A\ref{as:exact_memorization}: Verbatim Memorization}. We expand this assumption from the instance level to the token level, assuming an LM will memorize some tokens in seen instances. We measure the percentage of tokens in an instance ranked as the $k$ most likely in casual language modeling (\textbf{Mem $k$}). The $k$ value of 1 represents greedy decoding, and larger than 1 simulates the decoding with top $k$ token sampling.

\textbf{Assumption A\ref{as:Generation_Variation}: Generation Variation}. We evaluate the assumption that, given a seen prefix sequence, the LM exhibits less variation (i.e., higher certainty) in predicting the next token under different token sampling strategies. Since lower entropy indicates greater certainty, we measure entropy over the top $k$ most likely tokens (\textbf{Entropy $k$}) (see Appendix \ref{appd:entropy} for details).

\subsection{Experiment Design}\label{sec:case_experiment_design}

To enhance the generalizability of our results, we evaluate these assumptions using four different types of LLMs, with datasets used in different training phases, shown in Table \ref{tab:models_and_datasets}. We also investigate the impact of model size on MIA performance, using seven Pythia models with parameter sizes ranging from 70M to 13B.

Except for the UltraChat dataset, each dataset consists of multiple smaller subsets from different domains. For our experiments, we randomly sample 9 subsets from each, and consider each subset as an individual dataset.  To minimize distribution shifts between seen and unseen datasets, we randomly select 1,000 instances from the training split (seen) and 1,000 instances from the test split (unseen) within each dataset. If a test split is unavailable, we sample from the validation split. If there are fewer than 1,000 unseen instances, we use the entire test split, ensuring that each split contains at least 100 instances. 

Following prior work \cite{shi2023detecting}, we evaluate MIA performance using the area under the ROC curve (AUC) at the instance level, representing the probability that a seen instance has a better score (higher or lower) than an unseen instance \cite{AUC}. 

\subsection{Results}
\subsubsection{Within-Domain MIA}
Table \ref{tab:overall} shows the AUC for each MIA method across representative subsets (domains) of each dataset. Complete results for all datasets and domains are available in Appendix \ref{sec:appendix_case_study}.

On pretraining datasets, all metrics perform close to random guessing, with AUC close to 50. We also observed the same pattern with different sizes of Pythia LMs. 
Our results for Assumption A\ref{as:prob_absolute} are consistent with critiques they received (see Section \ref{sec:abs_prob}). 
We suspect that during pretraining, LMs are more likely to learn underlying data distributions, instead of memorizing specific instances.

On fine-tuning datasets, we observed a great variation in MIA AUC across domains. The best-performing metric, PPL\_200 on the RE-2012temp dataset, can have an AUC as high as 99.4. This suggests that data contamination from memorizing training instances remains a risk during the fine-tuning phase. Overall, the performance of the perplexity-based metric improves as the number of tokens increases. This trend is linked to the fine-tuning process, where tokens at the beginning of the training instance serve as input prompts but are not explicitly learned during training.


\subsubsection{Cross-Domains MIA with Data Distribution Shifts}

Within the same domain, the similar average PPL between seen and unseen instances indicates that they have similar underlying distributions, but also a high variation (STD). However, PPL differs a lot across domains. We therefore examine the impact of distribution shifts from different domains on the MIA performance, with the scenario where seen and unseen instances are from different domains. 


\begin{figure}[h]
  \includegraphics[width=0.5\textwidth]{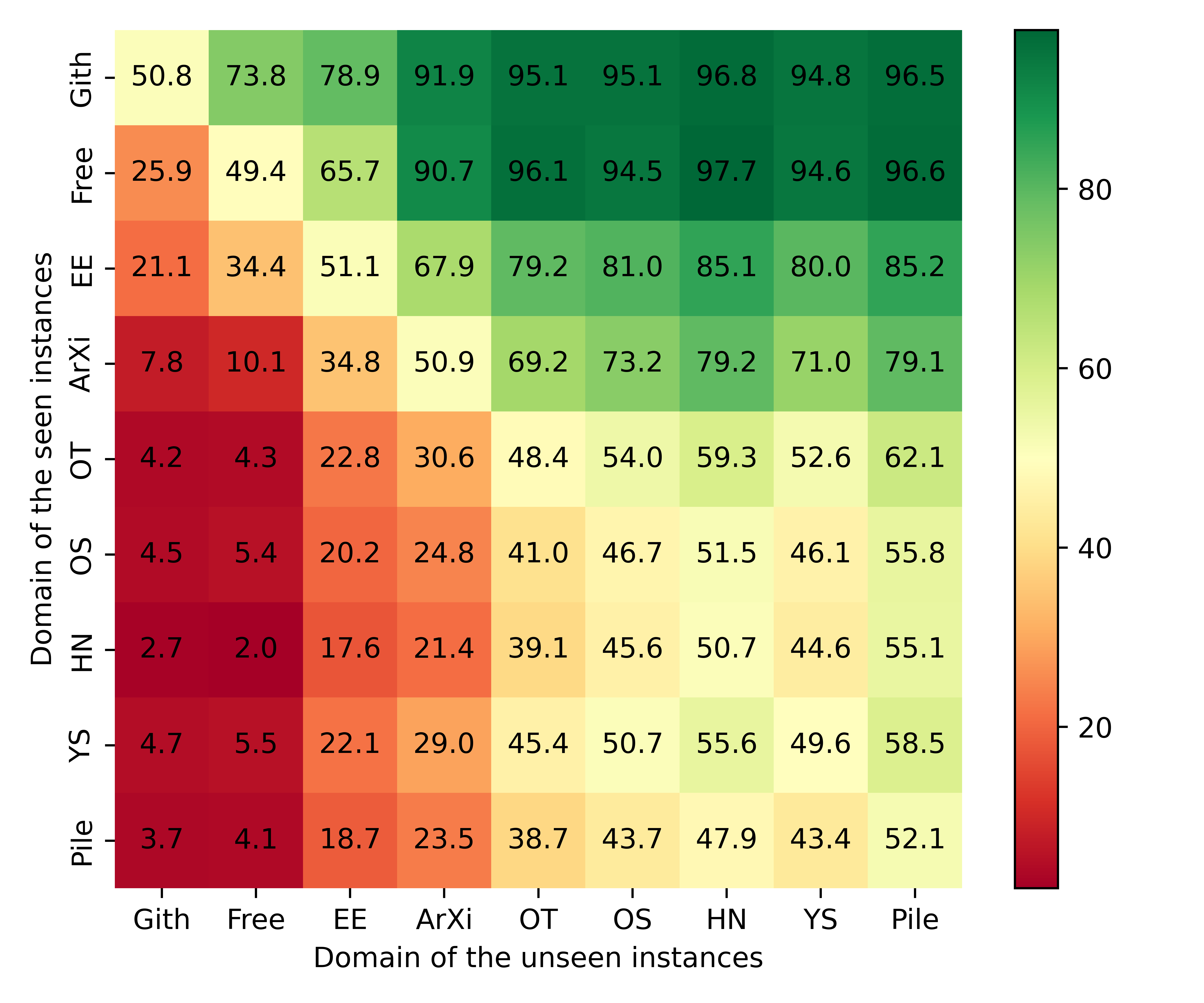}
  \vspace{-15pt}
  \caption{Average MIA AUC for the Pythia-6.9b model with PPL\_200, when the seen and unseen instances are from different domains. The abbreviations represent the domains in Table \ref{tab:results_pythia-6.9b} in the Appendix \ref{sec:appendix_case_study}. 
  }
  \label{fig:69b_ppl_200}
  \vspace{-15pt}
\end{figure}

We present the AUC with PPL\_200 in Figure \ref{fig:69b_ppl_200}. 
The AUC is high in the top-right corner when seen instances are from a domain with lower average PPLs and unseen instances are from a domain with higher average PPLs. Conversely, the bottom-left corner has low AUCs.
This indicates that the accuracy of PPL\_200-based MIA highly depends on the domain difference, instead of the seen vs. unseen distinction.
More information about the PPL\_200 distribution within and across domains is in Appendix \ref{sec:appendix_distribution}.
A similar trend is observed with other metrics (see Appendix \ref{sec:appendix_distribution} \& \ref{sec:appendix_diff_domains}).

\section{Discussion}
In our case study, we observed the evaluated MIA approaches perform well only on certain domains of datasets used during the fine-tuning phase, but not during the pretraining phase. This discrepancy may be attributed to the significantly larger dataset 
 and batch sizes employed in the pretraining phase.

Together, our case study and prior research show that 6 out of the 8 assumptions listed in Table \ref{tab:summmary} can often be invalid under certain conditions. The other two unverified assumptions, key information memorization (A\ref{as:key_memorization}) and answer change due to memorization (A\ref{as:same_task}), depend on very specific requirements, complicating their evaluation. Overall, our findings suggest that MIA remains a challenging task. 

The limited effectiveness of MIA in pretraining phases suggests detecting data contamination in benchmarks remains an important challenge for LLM evaluation. While poor MIA performance may indicate a lack of direct instance memorization, models could still learn underlying distributional patterns of benchmark data, enabling artificially high performance on the benchmark datasets \cite{dekoninck2024evading}. On the other hand, privacy and copyright risks persist, as LLMs might learn from sensitive/proprietary data (e.g., patented concepts)  without triggering MIA alarms.

\section{Conclusion}
In this study, we present a comprehensive survey of 50 studies focused on data contamination detection and their underlying assumptions. Our theoretical analysis reveals that these assumptions may not apply consistently across different contexts. 

Our case studies showed that 3 out of the 8 assumptions are not universally applicable across all training phases and dataset domains, especially for datasets used in the pretraining stage. 
Our cross-domain MIA experiments additionally show that many assumptions measure an LM's goodness of fit, which is not necessarily the result of instance memorization due to data contamination. Thus, detecting data contamination remains challenging.
 
 


\vspace{20pt}


\section{Limitations}
In this study, we reviewed 50 papers on data contamination detection, but there may be additional relevant studies not captured by our collection methods. We primarily focus on data contamination detection approaches for English LMs. Other detection approaches for non-English LMs, data modalities beyond text, and other machine learning techniques may exist and could potentially be transferable to English LMs.

Several factors influence an LM's ability to memorize an instance. The relationship between MIA performance and the training dataset (e.g., domain, size, batch size), as observed in our case study, may vary for other LMs and their respective datasets.

\section{Ethical Considerations}
In this work, we employed multiple LMs and their training sets, which may contain sensitive and/or identifiable information. For example, the Pile \cite{gao2020Pile} includes content crawled from the Internet. The BioMistral-NLU-7B and its training set contain datasets derived from de-identified clinical notes \cite{fu2024biomistral}. Therefore, we only downloaded the necessary instances and published the numerical results from our experiments. In our project GitHub, we only release our code for data sampling and MIA approaches to ensure reproducibility; we do not publish any actual data instances or LMs. We recommend the community to check the corresponding regulations before deploying the datasets and LMs for other purposes.

\section{Acknowledgements}
This work was supported by the National Institutes of Health (NIH)—National Cancer Institute (Grant Nos. 1R01CA248422-01A1), National Library of Medicine (Grant No. 2R15LM01320902A1), and National Center for Advancing Translational Sciences of the National Institutes of Health (Grant No. UL1TR002319).  The content is solely the responsibility of the authors and does not necessarily represent the official views of the NIH.

We thank Chenxi Li, Tianmai Zhang, and the anonymous reviewers for their feedback during the paper revision.

\bibliography{custom}


\vspace{0.2in}
\appendix
\section{Appendix}
\subsection{Table of Notations}
We present the notations used in this manuscript in Table \ref{tab:notations}.
\begin{table}[h]
\centering
\resizebox{0.45\textwidth}{!}{
\begin{tabular}{ll}
\hline
\textbf{Notation} & \textbf{Definition} \\ \hline
 $x$& A language instance, as a sequence of tokens.  \\ \hline
 $(x_p,x_s)$& \begin{tabular}[l]{@{}l@{}}A prefix-suffix instance pair, which is a \\  common data format in natural \\ language generation (NLG) tasks. \end{tabular} \\\hline
  $(x_c,x_k)$& \begin{tabular}[l]{@{}l@{}}A context-key instance pair from slot-filling\\ tasks, as defined in Requirement R\ref{condi:key_info}. \end{tabular} \\\hline
 $M$& A language model. \\ \hline
 $M_{\cdot}(x)$& \begin{tabular}[l]{@{}l@{}}$M$'s output respect to an input $x$, given \\ a decoding setup $\cdot$. If $\cdot$ is not specified, \\we consider it as a fixed, but unknown \\ decoding state. \end{tabular}  \\\hline
 $D$& A dataset, as a set of language instances.\\\hline
 $D_M$ & $M$'s training set.\\\hline
 $b(x,x')$ & \begin{tabular}[l]{@{}l@{}} Binary indicator function for instance-level \\ contamination, which takes two instances \\ $x$ and $x'$ as inputs, and returns \textit{False} (0) or \\ \textit{True} (1), based on the instance similarity.\end{tabular} \\\hline
$S(x,x')$ & \begin{tabular}[l]{@{}l@{}} A function accessing the similarity between \\ two instances, $x$ and $x'$ and outputs \\  a real value. \end{tabular} \\\hline
 $f(M,x)$ & \begin{tabular}[l]{@{}l@{}} Gold standard for instance-level \\ contamination, as defined in Equation \ref{eq:instance_contamination}. \end{tabular} \\\hline
 $P_M(x)$ & \begin{tabular}[l]{@{}l@{}}The probability of the instance $x$ given \\ an LM $M$. \end{tabular}  \\\hline

\begin{tabular}[l]{@{}l@{}}Min $p$\% \\ Token\end{tabular} & \begin{tabular}[l]{@{}l@{}}The average probabilities of top $p\%$ \\ least likely tokens in an instance $x$, based   \\on a given LM $M$. \end{tabular}  \\\hline
$\tau$ & \begin{tabular}[l]{@{}l@{}}The contamination threshold for functions. \end{tabular}  \\\hline
$\text{Var}(\{M_{\cdot}(x_p)\})$ & \begin{tabular}[l]{@{}l@{}}The measure of variations of outputs \\ produced by $M$ under diverse, different \\sampling  strategies. given $x_p$\end{tabular}  \\\hline
PPL\_$k$ & \begin{tabular}[l]{@{}l@{}} The perplexity from an instance's first \\ $k$ tokens. \end{tabular} \\\hline
Entropy $k$& \begin{tabular}[l]{@{}l@{}}The entropy of the top $k$ most likely tokens \\for the next position,  defined by Equation \ref{eq:entropy}.\end{tabular}  \\\hline
$\text{Eval}(M,D)$ & \begin{tabular}[l]{@{}l@{}}The evaluation result of an LM $M$ on a \\ dataset $D$.\end{tabular}  \\\hline

\end{tabular}
}
\caption{Table of notations.}
\label{tab:notations}
\end{table}
\subsection{Risks and Mitigation Approaches for Data Contamination} \label{sec:appendix_risks_mitigation}
During our paper collection, we identify the relevant research on the risks and mitigation strategies for data contamination, which does not involve proposing or evaluating existing detection approaches for data contamination. While excluding those papers from the main text of this manuscript, we provide their citations in Table \ref{tab:risks_mitigation} in this Appendix.

\begin{table*}[]
\centering
\resizebox{\textwidth}{!}{
\begin{tabular}{ll}
\hline
\textbf{Citation} & \textbf{Content} \\ \hline
\citet{zhou2023don} & Impact of direct data contamination on test performance \\
\citet{dekoninck2024evading} & Impact of indirect data contamination on test performance \\
\citet{jacovi-etal-2023-stop} & Strategies to prevent contamination in benchmark datasets \\
\citet{zhu-etal-2024-clean} & Strategies to mitigate contamination in benchmark datasets \\
\citet{haimes2024benchmark} &  Mitigating data contamination in benchmarks through retrospectively creating held-out datasets. \\
\citet{Kusa2024leveraging} & Proposing an evaluation pipeline in Systematic Literature Review to mitigate data contamination. \\
\citet{mccoy2023much} & Evaluating the novelty of LM-generated text \\
\citet{tofu2024} & Studying unlearning methods to make LMs forget specific training data \\
\citet{bowen2024scalinglawsdatapoisoning} & Studying contributing factors behind data poisoning, with corrupted or malicious training data. \\
\citet{mitchell2023detectgpt} & Differentiates human vs. machine-generated text using probability curvature \\ \hline
\end{tabular}
}
\caption{Seleted relevant work to risks and mitigation approaches for data contamination.}
\label{tab:risks_mitigation}
\end{table*}


\subsection{Case Study for Instance Similarity} \label{sec:appendix:conditions}
To assess the applicability of instance similarity-based detection approaches (see Section \ref{sec_instance_similarity}), we analyzed how frequently their requirements, R\ref{condi:disclose} and R\ref{condi:accessible}, are met. We reviewed the top 10 models from the Vellum LLM leaderboard\footnote{\url{https://www.vellum.ai/llm-leaderboard\#model-comparison}. Accessed on Oct 6, 2024.}. As demonstrated in Table \ref{tab:condition12}, none of the models fulfilled R\ref{condi:disclose}, the most basic requirement. However, some models disclose their cut-off date for collecting training data \cite{achiam2023gpt,gpt35}.

\begin{table}[H]
\centering
\resizebox{0.45\textwidth}{!}{
\begin{tabular}{llcc}
\hline
\multirow{2}{*}{\textbf{Model}} & \multirow{2}{*}{\textbf{Citation}} & \multicolumn{2}{c}{\textbf{Meet Require.}} \\ \cline{3-4} 
 &  & R\ref{condi:disclose} & R\ref{condi:accessible} \\ \hline
Claude 3.5 Sonnet & \citet{Claude35Sonnet}& No & - \\
Claude 3 Opus & \citet{Claude3Opus}& No & - \\
Gemini 1.5 Pro & \citet{Gemini15Pro}& No & - \\
GPT-4 & \citet{achiam2023gpt}& No & - \\
Llama 3 Instruct - 70B & \citet{llama3}& No & - \\
Claude 3 Haiku & \citet{Claude3Haiku}& No & - \\
GPT-3.5 & \citet{gpt35}& No & - \\
Mixtral 8x7B & \citet{jiang2024mixtral}& No & - \\
GPT-4o & \citet{gpt4o}& No & - \\
GPT-4o mini & \citet{gpt4omini}& No & - \\ \hline
\end{tabular}
}
\caption{None of the top 10 LMs, in the LLM Leaderboard by Vellum meet the requirements of disclosing pre-training corpora (R\ref{condi:disclose}).}
\label{tab:condition12}
\end{table}

\subsection{Entropy Calculation}\label{appd:entropy}
In this section, we explain the procedure for verifying Assumption A\ref{as:Generation_Variation}. In the context of casual language modeling, we consider an LM $M$ with a given prefix sequence $x_p$. Assumption A\ref{as:Generation_Variation} assumes that given $x_p$, if $M$ has seen an instance with the same prefix, it will generate similar responses, regardless of the sampling strategy used. Since the verification of this assumption can be influenced by various sampling strategies, we quantify the variance in the model's output by measuring the entropy of the token probabilities, which indicates the model’s certainty about the next token generation. 

To do this, we first compute the probability distribution of the next token over the model's vocabulary. Given that LLMs may contain vocabularies with over 50,000 tokens \cite{biderman2023pythia}, most tokens have a very low likelihood of being sampled. Therefore, we focus on the Entropy among the top $k$ most likely tokens (\textbf{Entropy $k$}).

At every token position $x_p$, we calculate the entropy based on the probabilities of the top $k$ tokens, using the following formula:
 \begin{equation}\label{eq:entropy}
 \begin{split}
 \text{Entropy}_k&(M,x_p)\\&=-\sum_{i=1}^{k}P_i(M(x_p)) \log P_i(M(x_p))
 \end{split}
 \end{equation}

Given an instance $x$ with $N$ tokens, the Entropy $k$ for $x$ is the average $\text{Entropy}_k(M,x_p)$ across all tokens $x_p$ in $x$:
 
 \begin{equation}\label{eq:entropy_avg}
 \text{Entropy}_k(M,x)=(\sum_{p=1}^{N}\text{Entropy}_k(M,x_p))/N
 \end{equation}
 



\subsection{More Case Study Results}

\subsubsection{Within-Domain Detection with Different LMs}\label{sec:appendix_case_study}

In this section, we present the detailed detection AUCs for all models: (1) different sizes of Pythia Models: Pythia-70m (Table \ref{tab:results_pythia-70m}), Pythia-160m (Table \ref{tab:results_pythia-160m}), Pythia-410m (Table \ref{tab:results_pythia-410m}),  Pythia-1.4b (Table \ref{tab:results_pythia-1.4b}), Pythia-2.8b (Table \ref{tab:results_pythia-2.8b}), Pythia-6.9b (Table \ref{tab:results_pythia-6.9b}), Pythia-12b (Table \ref{tab:results_pythia-12b}); (2) OLMo-2-7B (Table \ref{tab:results_OLMo-2-1124-7B}); and (3) BioMistral-NLU-7B (Table \ref{tab:results_BioMistral}).

Similar to the results in Table \ref{tab:overall}, we observed close-to-random performance in the detection AUCs for all Pythia models and dataset domains.

\begin{table*}[h]
\centering
\resizebox{0.98\textwidth}{!}{
\begin{tabular}{llccccccccccc}
\hline
\multicolumn{2}{l}{\textbf{Assumptions \& Metric}} 
& \textbf{Github}& \textbf{FreeLaw}& \textbf{\begin{tabular}[c]{@{}c@{}}Enron-\\ Emails\end{tabular}}& \textbf{ArXiv}& \textbf{\begin{tabular}[c]{@{}c@{}}OpenWeb-\\ Text2\end{tabular}}& \textbf{\begin{tabular}[c]{@{}c@{}}Open-\\ Subtitles\end{tabular}}& \textbf{\begin{tabular}[c]{@{}c@{}}Hacker-\\ News\end{tabular}}& \textbf{\begin{tabular}[c]{@{}c@{}}Youtube-\\ Subtitles\end{tabular}}& \textbf{Pile-CC}\\\hline
\multirow{6}{*}{A\ref{as:prob_absolute}}  & PPL\_50& 49.4 & 49.3 & 50.5 & 51.3 & 51.7 & 47.4 & 48.9 & 52.2 & 49.9 \\ 
 & PPL\_100& 50.3 & 50.2 & 51.5 & 50.8 & 50.5 & 46.2 & 48.1 & 52.0 & 50.8 \\ 
 & PPL\_200& 50.3 & 50.1 & 53.5 & 50.5 & 49.9 & 44.3 & 51.1 & 50.8 & 51.5 \\ 
 & Min 5\% token& 51.7 & 49.9 & 49.9 & 50.6 & 48.9 & 45.0 & 50.4 & 49.9 & 51.1 \\ 
 & Min 15\% token& 51.5 & 49.5 & 51.0 & 50.5 & 49.5 & 45.4 & 50.3 & 49.6 & 51.6 \\ 
 & Min 25\% token& 51.4 & 50.1 & 51.0 & 50.9 & 49.9 & 45.5 & 49.5 & 49.8 & 51.3 \\ 
 \cline{1-2}\multirow{3}{*}{A\ref{as:exact_memorization}} & Mem 5& 47.6 & 49.2 & 49.5 & 51.5 & 50.8 & 48.8 & 49.5 & 50.0 & 51.2 \\ 
 & Mem 15& 49.6 & 49.5 & 48.7 & 50.1 & 50.4 & 49.0 & 49.3 & 50.4 & 51.2 \\ 
 & Mem 25& 49.6 & 49.9 & 48.3 & 50.6 & 50.3 & 48.2 & 49.6 & 49.6 & 51.0 \\ 
\cline{1-2} \multirow{3}{*}{A\ref{as:Generation_Variation}} & Entropy 5& 50.8 & 49.7 & 48.2 & 52.1 & 49.9 & 47.4 & 49.3 & 50.4 & 50.2 \\ 
 & Entropy 15& 50.6 & 49.8 & 48.9 & 52.4 & 49.4 & 47.5 & 49.3 & 50.5 & 50.3 \\ 
 & Entropy 25& 50.6 & 49.9 & 49.2 & 52.4 & 49.0 & 47.7 & 49.3 & 50.2 & 50.2 \\\hline
\multicolumn{2}{l}{\textbf{Average AUC}}& 50.1 & 49.7 & 49.8 & 51.2 & 50.1 & 47.3 & 49.4 & 50.3 & 50.9 \\ 
\hline\multirow{2}{*}{PPL\_200} & Seen& 11.1$\pm$12.7 & 13.1$\pm$8.9 & 29.5$\pm$21.4 & 28.9$\pm$13.5 & 46.0$\pm$23.5 & 36.7$\pm$21.1 & 42.9$\pm$16.7 & 45.3$\pm$41.3 & 50.2$\pm$33.7 \\ 
 & Unseen& 11.5$\pm$20.9 & 14.2$\pm$25.3 & 31.7$\pm$22.4 & 29.0$\pm$13.1 & 45.4$\pm$22.5 & 33.7$\pm$18.2 & 43.7$\pm$18.8 & 44.3$\pm$28.7 & 51.5$\pm$35.6 \\ 

\hline
\end{tabular}
}
\caption{Average contamination detection AUC for the \textbf{pythia-70m} model,  under different domains within
the Pile dataset. `PPL\_200' represents the average perplexity $\pm$ STD, from the first 200 tokens within every instance.  The color {\color[HTML]{38761D} green} represents AUCs higher than 60.
}
\label{tab:results_pythia-70m}
\end{table*}

\begin{table*}[h]
\centering
\resizebox{0.98\textwidth}{!}{
\begin{tabular}{llccccccccccc}
\hline
\multicolumn{2}{l}{\textbf{Assumptions \& Metric}} 
& \textbf{Github}& \textbf{FreeLaw}& \textbf{\begin{tabular}[c]{@{}c@{}}Enron-\\ Emails\end{tabular}}& \textbf{ArXiv}& \textbf{\begin{tabular}[c]{@{}c@{}}OpenWeb-\\ Text2\end{tabular}}& \textbf{\begin{tabular}[c]{@{}c@{}}Open-\\ Subtitles\end{tabular}}& \textbf{\begin{tabular}[c]{@{}c@{}}Hacker-\\ News\end{tabular}}& \textbf{\begin{tabular}[c]{@{}c@{}}Youtube-\\ Subtitles\end{tabular}}& \textbf{Pile-CC}\\\hline
\multirow{6}{*}{A\ref{as:prob_absolute}}  & PPL\_50& 49.7 & 49.0 & 50.3 & 51.2 & 51.9 & 47.6 & 48.6 & 52.3 & 50.0 \\ 
 & PPL\_100& 50.3 & 49.4 & 51.2 & 51.0 & 50.4 & 46.1 & 47.6 & 51.7 & 50.7 \\ 
 & PPL\_200& 50.1 & 49.4 & 53.0 & 50.7 & 49.6 & 44.2 & 50.5 & 50.7 & 51.7 \\ 
 & Min 5\% token& 51.9 & 49.9 & 51.0 & 50.5 & 48.6 & 45.0 & 48.6 & 49.4 & 51.3 \\ 
 & Min 15\% token& 51.4 & 49.8 & 50.9 & 50.8 & 49.5 & 45.1 & 50.1 & 49.3 & 51.4 \\ 
 & Min 25\% token& 51.3 & 49.7 & 51.3 & 51.3 & 49.6 & 45.6 & 49.4 & 49.2 & 51.4 \\ 
 \cline{1-2}\multirow{3}{*}{A\ref{as:exact_memorization}} & Mem 5& 48.3 & 49.1 & 50.4 & 52.8 & 51.2 & 52.6 & 50.1 & 49.9 & 51.1 \\ 
 & Mem 15& 48.2 & 49.2 & 50.1 & 51.4 & 51.5 & 48.0 & 49.1 & 49.3 & 50.0 \\ 
 & Mem 25& 48.4 & 49.0 & 49.9 & 50.5 & 51.4 & 46.2 & 49.2 & 49.4 & 50.2 \\ 
\cline{1-2} \multirow{3}{*}{A\ref{as:Generation_Variation}} & Entropy 5& 51.3 & 49.1 & 48.9 & 52.1 & 49.8 & 47.0 & 48.8 & 50.1 & 50.6 \\ 
 & Entropy 15& 51.1 & 49.3 & 49.4 & 52.2 & 49.7 & 47.5 & 48.9 & 50.1 & 50.6 \\ 
 & Entropy 25& 51.0 & 49.4 & 49.6 & 52.1 & 49.4 & 47.7 & 48.9 & 50.0 & 50.6 \\\hline
\multicolumn{2}{l}{\textbf{Average AUC}}& 50.1 & 49.3 & 50.4 & 51.5 & 50.3 & 47.7 & 49.2 & 49.9 & 50.8 \\ 
\hline\multirow{2}{*}{PPL\_200} & Seen& 7.4$\pm$8.5 & 8.2$\pm$5.6 & 18.8$\pm$13.5 & 18.6$\pm$8.6 & 29.9$\pm$31.5 & 45.8$\pm$598.9 & 29.0$\pm$11.0 & 30.9$\pm$28.2 & 33.3$\pm$25.3 \\ 
 & Unseen& 7.6$\pm$13.5 & 8.8$\pm$16.7 & 20.3$\pm$14.8 & 18.7$\pm$8.3 & 28.5$\pm$12.8 & 24.8$\pm$11.4 & 29.6$\pm$13.1 & 30.4$\pm$23.0 & 34.4$\pm$25.9 \\ 

\hline
\end{tabular}
}
\caption{Average contamination detection AUC for the \textbf{pythia-160m} model,  under different domains within
the Pile dataset. `PPL\_200' represents the average perplexity $\pm$ STD, from the first 200 tokens within every instance.  The color {\color[HTML]{38761D} green} represents AUCs higher than 60.
}
\label{tab:results_pythia-160m}
\end{table*}

\begin{table*}[h]
\centering
\resizebox{0.98\textwidth}{!}{
\begin{tabular}{llccccccccccc}
\hline
\multicolumn{2}{l}{\textbf{Assumptions \& Metric}} 
& \textbf{Github}& \textbf{FreeLaw}& \textbf{\begin{tabular}[c]{@{}c@{}}Enron-\\ Emails\end{tabular}}& \textbf{ArXiv}& \textbf{\begin{tabular}[c]{@{}c@{}}OpenWeb-\\ Text2\end{tabular}}& \textbf{\begin{tabular}[c]{@{}c@{}}Open-\\ Subtitles\end{tabular}}& \textbf{\begin{tabular}[c]{@{}c@{}}Hacker-\\ News\end{tabular}}& \textbf{\begin{tabular}[c]{@{}c@{}}Youtube-\\ Subtitles\end{tabular}}& \textbf{Pile-CC}\\\hline
\multirow{6}{*}{A\ref{as:prob_absolute}}  & PPL\_50& 49.5 & 49.5 & 50.1 & 50.7 & 51.0 & 47.4 & 48.8 & 51.5 & 50.0 \\ 
 & PPL\_100& 50.5 & 49.2 & 51.3 & 50.9 & 49.7 & 45.5 & 47.6 & 51.0 & 50.3 \\ 
 & PPL\_200& 50.3 & 49.3 & 53.0 & 50.6 & 48.8 & 44.6 & 50.9 & 50.2 & 52.0 \\ 
 & Min 5\% token& 52.2 & 49.8 & 50.3 & 50.7 & 49.0 & 45.5 & 49.3 & 48.9 & 51.4 \\ 
 & Min 15\% token& 51.6 & 49.6 & 50.9 & 51.0 & 49.3 & 46.0 & 50.2 & 49.0 & 51.2 \\ 
 & Min 25\% token& 51.4 & 49.4 & 51.0 & 51.3 & 48.9 & 46.4 & 50.2 & 48.9 & 51.0 \\ 
 \cline{1-2}\multirow{3}{*}{A\ref{as:exact_memorization}} & Mem 5& 47.5 & 49.2 & 51.3 & 53.0 & 50.5 & 50.3 & 49.4 & 49.7 & 50.3 \\ 
 & Mem 15& 47.6 & 49.2 & 51.3 & 52.6 & 51.8 & 50.3 & 49.7 & 50.1 & 50.4 \\ 
 & Mem 25& 48.2 & 49.2 & 51.1 & 51.2 & 51.9 & 52.6 & 50.5 & 49.2 & 51.2 \\ 
\cline{1-2} \multirow{3}{*}{A\ref{as:Generation_Variation}} & Entropy 5& 51.2 & 49.1 & 49.4 & 52.2 & 50.5 & 47.4 & 49.0 & 49.5 & 50.2 \\ 
 & Entropy 15& 51.0 & 49.3 & 49.9 & 52.0 & 49.7 & 48.0 & 48.9 & 49.5 & 50.2 \\ 
 & Entropy 25& 50.9 & 49.4 & 50.1 & 51.9 & 49.4 & 48.2 & 48.9 & 49.2 & 50.3 \\\hline
\multicolumn{2}{l}{\textbf{Average AUC}}& 50.0 & 49.4 & 50.8 & 51.9 & 50.1 & 48.4 & 49.5 & 49.7 & 50.7 \\ 
\hline\multirow{2}{*}{PPL\_200} & Seen& 4.7$\pm$5.0 & 5.5$\pm$3.4 & 11.8$\pm$8.2 & 12.7$\pm$6.0 & 18.8$\pm$9.7 & 20.0$\pm$9.1 & 19.8$\pm$7.5 & 20.6$\pm$22.2 & 22.3$\pm$17.0 \\ 
 & Unseen& 4.8$\pm$7.2 & 5.9$\pm$11.4 & 12.9$\pm$9.6 & 12.7$\pm$5.7 & 18.2$\pm$8.3 & 18.5$\pm$7.9 & 20.4$\pm$9.3 & 20.0$\pm$14.8 & 23.0$\pm$17.8 \\ 

\hline
\end{tabular}
}
\caption{Average contamination detection AUC for the \textbf{pythia-410m} model,  under different domains within
the Pile dataset. `PPL\_200' represents the average perplexity $\pm$ STD, from the first 200 tokens within every instance.  The color {\color[HTML]{38761D} green} represents AUCs higher than 60.
}
\label{tab:results_pythia-410m}
\end{table*}

\begin{table*}[h]
\centering
\resizebox{0.98\textwidth}{!}{
\begin{tabular}{llccccccccccc}
\hline
\multicolumn{2}{l}{\textbf{Assumptions \& Metric}} 
& \textbf{Github}& \textbf{FreeLaw}& \textbf{\begin{tabular}[c]{@{}c@{}}Enron-\\ Emails\end{tabular}}& \textbf{ArXiv}& \textbf{\begin{tabular}[c]{@{}c@{}}OpenWeb-\\ Text2\end{tabular}}& \textbf{\begin{tabular}[c]{@{}c@{}}Open-\\ Subtitles\end{tabular}}& \textbf{\begin{tabular}[c]{@{}c@{}}Hacker-\\ News\end{tabular}}& \textbf{\begin{tabular}[c]{@{}c@{}}Youtube-\\ Subtitles\end{tabular}}& \textbf{Pile-CC}\\\hline
\multirow{6}{*}{A\ref{as:prob_absolute}}  & PPL\_50& 49.5 & 49.0 & 49.8 & 50.7 & 50.7 & 48.1 & 48.5 & 51.0 & 49.8 \\ 
 & PPL\_100& 50.7 & 49.5 & 50.9 & 51.0 & 49.2 & 45.6 & 47.4 & 50.9 & 50.2 \\ 
 & PPL\_200& 50.8 & 49.8 & 51.9 & 50.9 & 48.7 & 44.4 & 50.9 & 49.9 & 52.0 \\ 
 & Min 5\% token& 51.8 & 50.5 & 49.7 & 51.0 & 49.2 & 46.2 & 48.9 & 48.4 & 51.1 \\ 
 & Min 15\% token& 51.5 & 49.7 & 50.3 & 51.3 & 49.6 & 47.0 & 50.1 & 48.7 & 51.1 \\ 
 & Min 25\% token& 51.4 & 49.4 & 50.5 & 51.5 & 49.1 & 47.8 & 50.0 & 48.6 & 51.2 \\ 
 \cline{1-2}\multirow{3}{*}{A\ref{as:exact_memorization}} & Mem 5& 48.8 & 49.1 & 50.9 & 53.0 & 51.0 & 51.5 & 49.6 & 49.1 & 50.5 \\ 
 & Mem 15& 48.0 & 49.5 & 50.8 & 51.4 & 51.2 & 51.7 & 49.8 & 48.9 & 50.4 \\ 
 & Mem 25& 48.7 & 49.5 & 51.1 & 51.0 & 51.5 & 50.9 & 49.7 & 48.7 & 51.0 \\ 
\cline{1-2} \multirow{3}{*}{A\ref{as:Generation_Variation}} & Entropy 5& 51.2 & 49.0 & 49.9 & 52.0 & 49.9 & 46.6 & 48.7 & 49.3 & 50.2 \\ 
 & Entropy 15& 51.0 & 49.1 & 50.1 & 51.9 & 49.5 & 48.0 & 48.7 & 48.9 & 50.2 \\ 
 & Entropy 25& 51.0 & 49.2 & 50.1 & 51.8 & 49.3 & 48.5 & 48.8 & 48.7 & 50.3 \\\hline
\multicolumn{2}{l}{\textbf{Average AUC}}& 50.2 & 49.4 & 50.5 & 51.7 & 50.1 & 48.6 & 49.2 & 49.1 & 50.6 \\ 
\hline\multirow{2}{*}{PPL\_200} & Seen& 3.6$\pm$3.8 & 4.4$\pm$2.4 & 8.5$\pm$5.9 & 9.8$\pm$4.6 & 14.2$\pm$7.4 & 16.0$\pm$6.9 & 15.3$\pm$5.7 & 16.3$\pm$19.2 & 17.1$\pm$12.5 \\ 
 & Unseen& 3.6$\pm$4.9 & 4.7$\pm$8.6 & 9.3$\pm$7.3 & 9.9$\pm$4.4 & 13.7$\pm$6.4 & 14.8$\pm$6.1 & 15.7$\pm$7.2 & 15.7$\pm$12.3 & 17.8$\pm$13.9 \\ 

\hline
\end{tabular}
}
\caption{Average contamination detection AUC for the \textbf{pythia-1.4b} model,  under different domains within
the Pile dataset. `PPL\_200' represents the average perplexity $\pm$ STD, from the first 200 tokens within every instance.  The color {\color[HTML]{38761D} green} represents AUCs higher than 60.
}
\label{tab:results_pythia-1.4b}
\end{table*}

\begin{table*}[h]
\centering
\resizebox{0.98\textwidth}{!}{
\begin{tabular}{llccccccccccc}
\hline
\multicolumn{2}{l}{\textbf{Assumptions \& Metric}} 
& \textbf{Github}& \textbf{FreeLaw}& \textbf{\begin{tabular}[c]{@{}c@{}}Enron-\\ Emails\end{tabular}}& \textbf{ArXiv}& \textbf{\begin{tabular}[c]{@{}c@{}}OpenWeb-\\ Text2\end{tabular}}& \textbf{\begin{tabular}[c]{@{}c@{}}Open-\\ Subtitles\end{tabular}}& \textbf{\begin{tabular}[c]{@{}c@{}}Hacker-\\ News\end{tabular}}& \textbf{\begin{tabular}[c]{@{}c@{}}Youtube-\\ Subtitles\end{tabular}}& \textbf{Pile-CC}\\\hline
\multirow{6}{*}{A\ref{as:prob_absolute}}  & PPL\_50& 49.4 & 48.6 & 49.4 & 50.7 & 50.5 & 47.8 & 48.0 & 51.1 & 49.9 \\ 
 & PPL\_100& 50.5 & 49.2 & 50.6 & 50.9 & 49.0 & 45.9 & 47.5 & 51.0 & 50.4 \\ 
 & PPL\_200& 50.6 & 49.6 & 51.7 & 50.8 & 48.6 & 45.3 & 50.8 & 49.9 & 52.2 \\ 
 & Min 5\% token& 51.6 & 49.8 & 49.9 & 51.2 & 49.5 & 47.2 & 48.2 & 48.6 & 51.6 \\ 
 & Min 15\% token& 51.5 & 49.6 & 50.3 & 51.4 & 49.5 & 48.1 & 49.4 & 48.7 & 51.3 \\ 
 & Min 25\% token& 51.2 & 49.4 & 50.1 & 51.4 & 48.9 & 48.8 & 49.7 & 48.6 & 51.0 \\ 
 \cline{1-2}\multirow{3}{*}{A\ref{as:exact_memorization}} & Mem 5& 48.7 & 49.1 & 50.7 & 52.8 & 50.9 & 51.5 & 49.5 & 50.2 & 50.9 \\ 
 & Mem 15& 48.2 & 48.9 & 50.4 & 51.0 & 51.3 & 50.4 & 48.9 & 49.7 & 50.2 \\ 
 & Mem 25& 48.9 & 48.5 & 50.5 & 50.8 & 51.4 & 50.0 & 48.7 & 49.3 & 50.2 \\ 
\cline{1-2} \multirow{3}{*}{A\ref{as:Generation_Variation}} & Entropy 5& 50.9 & 49.1 & 49.5 & 51.9 & 50.0 & 47.5 & 48.9 & 49.0 & 50.6 \\ 
 & Entropy 15& 50.8 & 49.2 & 49.8 & 51.9 & 49.5 & 48.7 & 49.0 & 48.9 & 50.5 \\ 
 & Entropy 25& 50.8 & 49.2 & 50.0 & 51.8 & 49.3 & 49.1 & 49.0 & 48.8 & 50.6 \\\hline
\multicolumn{2}{l}{\textbf{Average AUC}}& 50.1 & 49.2 & 50.2 & 51.6 & 50.0 & 48.8 & 49.0 & 49.4 & 50.8 \\ 
\hline\multirow{2}{*}{PPL\_200} & Seen& 3.2$\pm$3.3 & 3.9$\pm$2.1 & 7.2$\pm$5.1 & 8.7$\pm$4.1 & 12.5$\pm$6.5 & 14.0$\pm$6.2 & 13.3$\pm$4.9 & 14.4$\pm$17.3 & 15.0$\pm$10.2 \\ 
 & Unseen& 3.2$\pm$4.4 & 4.2$\pm$7.8 & 7.9$\pm$6.3 & 8.7$\pm$3.9 & 12.1$\pm$5.8 & 13.1$\pm$5.5 & 13.6$\pm$6.1 & 13.9$\pm$11.3 & 15.7$\pm$12.2 \\ 

\hline
\end{tabular}
}
\caption{Average contamination detection AUC for the \textbf{pythia-2.8b} model,  under different domains within
the Pile dataset. `PPL\_200' represents the average perplexity $\pm$ STD, from the first 200 tokens within every instance.  The color {\color[HTML]{38761D} green} represents AUCs higher than 60.
}
\label{tab:results_pythia-2.8b}
\end{table*}

\begin{table*}[h]
\centering
\resizebox{0.98\textwidth}{!}{
\begin{tabular}{llccccccccccc}
\hline
\multicolumn{2}{l}{\textbf{Assumptions \& Metric}} 
& \textbf{Github}& \textbf{FreeLaw}& \textbf{\begin{tabular}[c]{@{}c@{}}Enron-\\ Emails\end{tabular}}& \textbf{ArXiv}& \textbf{\begin{tabular}[c]{@{}c@{}}OpenWeb-\\ Text2\end{tabular}}& \textbf{\begin{tabular}[c]{@{}c@{}}Open-\\ Subtitles\end{tabular}}& \textbf{\begin{tabular}[c]{@{}c@{}}Hacker-\\ News\end{tabular}}& \textbf{\begin{tabular}[c]{@{}c@{}}Youtube-\\ Subtitles\end{tabular}}& \textbf{Pile-CC}\\\hline
\multirow{6}{*}{A\ref{as:prob_absolute}}  & PPL\_50& 49.7 & 48.9 & 49.5 & 50.7 & 50.3 & 47.7 & 48.5 & 50.7 & 49.8 \\ 
 & PPL\_100& 50.7 & 49.4 & 50.3 & 51.1 & 48.8 & 46.2 & 47.4 & 50.4 & 50.4 \\ 
 & PPL\_200& 50.8 & 49.4 & 51.1 & 50.9 & 48.4 & 46.7 & 50.7 & 49.6 & 52.1 \\ 
 & Min 5\% token& 51.7 & 49.8 & 49.2 & 51.3 & 49.7 & 48.2 & 47.8 & 48.5 & 50.9 \\ 
 & Min 15\% token& 51.6 & 49.7 & 49.8 & 51.3 & 49.1 & 49.5 & 49.7 & 48.6 & 51.1 \\ 
 & Min 25\% token& 51.3 & 49.3 & 50.0 & 51.4 & 48.8 & 50.2 & 49.8 & 48.5 & 51.1 \\ 
 \cline{1-2}\multirow{3}{*}{A\ref{as:exact_memorization}} & Mem 5& 49.2 & 48.7 & 50.6 & 52.7 & 50.5 & 50.0 & 49.9 & 49.1 & 50.8 \\ 
 & Mem 15& 48.9 & 48.9 & 51.0 & 51.9 & 51.8 & 50.6 & 49.2 & 48.6 & 50.7 \\ 
 & Mem 25& 49.6 & 48.8 & 51.1 & 51.6 & 51.1 & 50.3 & 49.5 & 48.2 & 51.3 \\ 
\cline{1-2} \multirow{3}{*}{A\ref{as:Generation_Variation}} & Entropy 5& 51.0 & 49.1 & 49.5 & 52.0 & 49.7 & 48.4 & 49.1 & 49.3 & 50.9 \\ 
 & Entropy 15& 50.8 & 49.1 & 49.7 & 52.0 & 49.3 & 49.6 & 49.0 & 49.0 & 50.9 \\ 
 & Entropy 25& 50.8 & 49.2 & 49.8 & 51.9 & 49.1 & 50.0 & 49.0 & 48.9 & 51.0 \\\hline
\multicolumn{2}{l}{\textbf{Average AUC}}& 50.3 & 49.2 & 50.2 & 51.8 & 49.9 & 49.4 & 49.3 & 49.0 & 50.9 \\ 
\hline\multirow{2}{*}{PPL\_200} & Seen& 2.8$\pm$3.0 & 3.6$\pm$1.9 & 6.1$\pm$4.4 & 7.9$\pm$3.7 & 11.2$\pm$5.8 & 12.2$\pm$5.9 & 12.2$\pm$4.5 & 13.2$\pm$16.0 & 13.7$\pm$9.1 \\ 
 & Unseen& 2.9$\pm$4.1 & 3.8$\pm$6.2 & 6.7$\pm$5.6 & 8.0$\pm$3.6 & 10.9$\pm$5.3 & 11.5$\pm$5.1 & 12.4$\pm$5.4 & 12.7$\pm$10.6 & 14.3$\pm$11.0 \\ 

\hline
\end{tabular}
}
\caption{Average contamination detection AUC for the \textbf{pythia-6.9b} model,  under different domains within
the Pile dataset. `PPL\_200' represents the average perplexity $\pm$ STD, from the first 200 tokens within every instance.  The color {\color[HTML]{38761D} green} represents AUCs higher than 60.
}
\label{tab:results_pythia-6.9b}
\end{table*}

\begin{table*}[h]
\centering
\resizebox{0.98\textwidth}{!}{
\begin{tabular}{llccccccccccc}
\hline
\multicolumn{2}{l}{\textbf{Assumptions \& Metric}} 
& \textbf{Github}& \textbf{FreeLaw}& \textbf{\begin{tabular}[c]{@{}c@{}}Enron-\\ Emails\end{tabular}}& \textbf{ArXiv}& \textbf{\begin{tabular}[c]{@{}c@{}}OpenWeb-\\ Text2\end{tabular}}& \textbf{\begin{tabular}[c]{@{}c@{}}Open-\\ Subtitles\end{tabular}}& \textbf{\begin{tabular}[c]{@{}c@{}}Hacker-\\ News\end{tabular}}& \textbf{\begin{tabular}[c]{@{}c@{}}Youtube-\\ Subtitles\end{tabular}}& \textbf{Pile-CC}\\\hline
\multirow{6}{*}{A\ref{as:prob_absolute}}  & PPL\_50& 49.9 & 48.7 & 49.5 & 51.0 & 50.4 & 48.5 & 48.3 & 50.7 & 49.8 \\ 
 & PPL\_100& 50.9 & 49.6 & 50.2 & 50.9 & 48.9 & 47.4 & 47.3 & 50.2 & 50.3 \\ 
 & PPL\_200& 50.9 & 49.5 & 51.0 & 51.0 & 48.5 & 48.4 & 50.5 & 49.4 & 51.8 \\ 
 & Min 5\% token& 51.9 & 50.2 & 49.2 & 51.4 & 49.5 & 49.2 & 47.8 & 48.7 & 50.9 \\ 
 & Min 15\% token& 51.7 & 49.4 & 49.8 & 51.3 & 49.0 & 50.2 & 49.4 & 48.7 & 51.0 \\ 
 & Min 25\% token& 51.4 & 49.1 & 49.9 & 51.4 & 48.6 & 50.8 & 49.3 & 48.8 & 51.0 \\ 
 \cline{1-2}\multirow{3}{*}{A\ref{as:exact_memorization}} & Mem 5& 49.0 & 49.0 & 50.9 & 53.9 & 51.0 & 52.6 & 49.6 & 48.5 & 50.9 \\ 
 & Mem 15& 48.5 & 49.5 & 51.0 & 52.5 & 51.1 & 49.0 & 49.1 & 48.3 & 50.0 \\ 
 & Mem 25& 49.5 & 49.2 & 50.6 & 52.7 & 50.7 & 48.6 & 49.2 & 48.2 & 51.4 \\ 
\cline{1-2} \multirow{3}{*}{A\ref{as:Generation_Variation}} & Entropy 5& 51.0 & 49.0 & 49.5 & 52.0 & 50.0 & 49.2 & 48.8 & 49.3 & 51.0 \\ 
 & Entropy 15& 50.9 & 49.1 & 49.7 & 51.9 & 49.6 & 50.6 & 48.8 & 49.1 & 50.9 \\ 
 & Entropy 25& 50.9 & 49.2 & 49.8 & 51.8 & 49.4 & 51.0 & 48.8 & 49.0 & 50.9 \\\hline
\multicolumn{2}{l}{\textbf{Average AUC}}& 50.3 & 49.2 & 50.2 & 52.1 & 49.9 & 49.9 & 48.9 & 48.9 & 50.8 \\ 
\hline\multirow{2}{*}{PPL\_200} & Seen& 2.6$\pm$2.8 & 3.4$\pm$1.7 & 5.4$\pm$4.0 & 7.5$\pm$3.5 & 10.4$\pm$5.4 & 11.0$\pm$5.7 & 11.2$\pm$4.0 & 12.3$\pm$15.0 & 12.9$\pm$8.4 \\ 
 & Unseen& 2.7$\pm$3.7 & 3.6$\pm$5.6 & 5.9$\pm$5.0 & 7.5$\pm$3.4 & 10.1$\pm$5.0 & 10.6$\pm$5.0 & 11.5$\pm$5.0 & 11.8$\pm$10.0 & 13.4$\pm$10.1 \\ 

\hline
\end{tabular}
}
\caption{Average contamination detection AUC for the \textbf{pythia-12b} model,  under different domains within
the Pile dataset. `PPL\_200' represents the average perplexity $\pm$ STD, from the first 200 tokens within every instance.  The color {\color[HTML]{38761D} green} represents AUCs higher than 60.
}
\label{tab:results_pythia-12b}
\end{table*}

\begin{table*}[h]
\centering
\resizebox{0.98\textwidth}{!}{
\begin{tabular}{llccccccccccc}
\hline
\multicolumn{2}{l}{\textbf{Assumptions \& Metric}} 
& \textbf{cpp}& \textbf{python}& \textbf{Github-Lean}& \textbf{julia}& \textbf{tex}& \textbf{Github-Isabelle}& \textbf{fortran}& \textbf{Github-Coq}& \textbf{r}\\\hline
\multirow{6}{*}{A\ref{as:prob_absolute}}  & PPL\_50& 51.1 & 51.7 & 49.9 & 50.2 & 49.4 & 50.9 & 50.3 & 49.1 & 49.5 \\ 
 & PPL\_100& 51.4 & 51.6 & 52.0 & 51.3 & 50.4 & 51.7 & 50.6 & 49.1 & 51.4 \\ 
 & PPL\_200& 51.8 & 50.8 & 51.2 & 51.6 & 50.0 & 50.8 & 51.1 & 49.0 & 53.0 \\ 
 & Min 5\% token& 50.1 & 51.1 & 49.0 & 52.0 & 49.4 & 51.2 & 48.1 & 49.4 & 50.9 \\ 
 & Min 15\% token& 50.2 & 50.4 & 50.7 & 51.2 & 48.6 & 50.4 & 48.1 & 49.0 & 51.5 \\ 
 & Min 25\% token& 50.3 & 49.8 & 51.3 & 51.0 & 48.7 & 49.9 & 48.2 & 48.9 & 52.4 \\ 
 \cline{1-2}\multirow{3}{*}{A\ref{as:exact_memorization}} & Mem 5& 51.3 & 48.2 & 51.8 & 49.0 & 48.2 & 55.4 & 50.8 & 50.5 & 51.6 \\ 
 & Mem 15& 52.0 & 49.6 & 50.7 & 48.3 & 48.5 & 54.6 & 50.6 & 50.4 & 52.0 \\ 
 & Mem 25& 51.6 & 50.5 & 50.1 & 49.1 & 49.4 & 54.3 & 50.4 & 50.8 & 50.9 \\ 
\cline{1-2} \multirow{3}{*}{A\ref{as:Generation_Variation}} & Entropy 5& 50.2 & 49.4 & 51.7 & 51.3 & 48.8 & 49.0 & 48.8 & 48.9 & 54.1 \\ 
 & Entropy 15& 50.2 & 49.4 & 51.7 & 51.3 & 48.8 & 48.9 & 48.8 & 49.0 & 53.5 \\ 
 & Entropy 25& 50.3 & 49.5 & 51.8 & 51.2 & 48.8 & 49.0 & 48.8 & 49.0 & 53.5 \\\hline
\multicolumn{2}{l}{\textbf{Average AUC}}& 50.9 & 49.8 & 50.9 & 50.4 & 49.0 & 51.6 & 49.6 & 49.7 & 52.2 \\ 
\hline\multirow{2}{*}{PPL\_200} & Seen& 4.3$\pm$2.6 & 6.7$\pm$3.9 & 6.9$\pm$3.4 & 8.3$\pm$5.0 & 8.3$\pm$5.1 & 8.7$\pm$5.4 & 9.5$\pm$6.8 & 10.4$\pm$8.3 & 10.5$\pm$7.0 \\ 
 & Unseen& 4.6$\pm$3.0 & 6.8$\pm$4.1 & 6.9$\pm$3.0 & 8.6$\pm$5.6 & 8.7$\pm$6.3 & 9.2$\pm$6.5 & 9.9$\pm$7.6 & 9.9$\pm$7.2 & 10.5$\pm$5.6 \\ 

\hline
\end{tabular}
}
\caption{Average contamination detection AUC for the \textbf{OLMo-2-1124-7B} model,  under different domains within
the Algebraic Stack dataset. `PPL\_200' represents the average perplexity $\pm$ STD, from the first 200 tokens within every instance.  The color {\color[HTML]{38761D} green} represents AUCs higher than 60.
}
\label{tab:results_OLMo-2-1124-7B}
\end{table*}

\begin{table*}[h]
\centering
\resizebox{0.98\textwidth}{!}{
\begin{tabular}{llccccccccccc}
\hline
\multicolumn{2}{l}{\textbf{Assumptions \& Metric}} 
& \textbf{\begin{tabular}[c]{@{}c@{}}RE-\\ 2012temp\end{tabular}}& \textbf{\begin{tabular}[c]{@{}c@{}}STS-\\ B\end{tabular}}& \textbf{\begin{tabular}[c]{@{}c@{}}DC-\\ MTSample\end{tabular}}& \textbf{\begin{tabular}[c]{@{}c@{}}RE-\\ 2011coref\end{tabular}}& \textbf{\begin{tabular}[c]{@{}c@{}}events-\\ BioRed\end{tabular}}& \textbf{\begin{tabular}[c]{@{}c@{}}events-\\ NLMGene\end{tabular}}& \textbf{\begin{tabular}[c]{@{}c@{}}events-\\ 2012temp\end{tabular}}& \textbf{\begin{tabular}[c]{@{}c@{}}events-\\ 2006deid\end{tabular}}& \textbf{\begin{tabular}[c]{@{}c@{}}events-\\ BioASQ\end{tabular}}\\\hline
\multirow{6}{*}{A\ref{as:prob_absolute}}  & PPL\_50& {\color[HTML]{38761D} \textbf{62.5}} & {\color[HTML]{38761D} \textbf{83.3}} & 51.9 & {\color[HTML]{38761D} \textbf{60.5}} & 50.4 & 49.4 & 50.0 & 51.4 & {\color[HTML]{38761D} \textbf{67.6}} \\ 
 & PPL\_100& {\color[HTML]{38761D} \textbf{95.3}} & {\color[HTML]{38761D} \textbf{93.3}} & 60.0 & {\color[HTML]{38761D} \textbf{70.0}} & 50.0 & 48.8 & 50.4 & 59.3 & 52.6 \\ 
 & PPL\_200& {\color[HTML]{38761D} \textbf{99.4}} & {\color[HTML]{38761D} \textbf{96.8}} & 58.9 & {\color[HTML]{38761D} \textbf{70.5}} & {\color[HTML]{38761D} \textbf{89.4}} & {\color[HTML]{38761D} \textbf{87.4}} & {\color[HTML]{38761D} \textbf{76.2}} & {\color[HTML]{38761D} \textbf{79.0}} & {\color[HTML]{38761D} \textbf{66.1}} \\ 
 & Min 5\% token& {\color[HTML]{38761D} \textbf{92.9}} & {\color[HTML]{38761D} \textbf{93.4}} & 47.4 & {\color[HTML]{38761D} \textbf{72.6}} & {\color[HTML]{38761D} \textbf{61.4}} & {\color[HTML]{38761D} \textbf{76.1}} & 57.5 & {\color[HTML]{38761D} \textbf{78.0}} & {\color[HTML]{38761D} \textbf{74.9}} \\ 
 & Min 15\% token& {\color[HTML]{38761D} \textbf{93.3}} & {\color[HTML]{38761D} \textbf{93.4}} & 51.7 & {\color[HTML]{38761D} \textbf{70.1}} & {\color[HTML]{38761D} \textbf{88.3}} & {\color[HTML]{38761D} \textbf{85.3}} & {\color[HTML]{38761D} \textbf{71.2}} & {\color[HTML]{38761D} \textbf{82.3}} & {\color[HTML]{38761D} \textbf{69.4}} \\ 
 & Min 25\% token& {\color[HTML]{38761D} \textbf{93.4}} & {\color[HTML]{38761D} \textbf{92.1}} & 53.3 & {\color[HTML]{38761D} \textbf{68.5}} & {\color[HTML]{38761D} \textbf{94.1}} & {\color[HTML]{38761D} \textbf{80.5}} & {\color[HTML]{38761D} \textbf{74.8}} & {\color[HTML]{38761D} \textbf{76.5}} & {\color[HTML]{38761D} \textbf{64.7}} \\ 
 \cline{1-2}\multirow{3}{*}{A\ref{as:exact_memorization}} & Mem 5& 41.4 & 48.0 & 47.9 & 46.6 & 49.8 & 52.0 & 51.0 & 52.5 & 49.2 \\ 
 & Mem 15& 45.2 & 53.3 & 49.1 & 46.4 & 48.9 & 52.3 & 51.9 & 52.7 & 52.8 \\ 
 & Mem 25& 48.9 & 55.9 & 50.3 & 48.4 & 46.6 & 52.1 & 52.6 & 53.1 & 52.5 \\ 
\cline{1-2} \multirow{3}{*}{A\ref{as:Generation_Variation}} & Entropy 5& {\color[HTML]{38761D} \textbf{93.2}} & {\color[HTML]{38761D} \textbf{80.5}} & 55.4 & {\color[HTML]{38761D} \textbf{63.9}} & {\color[HTML]{38761D} \textbf{94.5}} & {\color[HTML]{38761D} \textbf{65.3}} & {\color[HTML]{38761D} \textbf{74.3}} & {\color[HTML]{38761D} \textbf{62.6}} & 43.1 \\ 
 & Entropy 15& {\color[HTML]{38761D} \textbf{93.1}} & {\color[HTML]{38761D} \textbf{82.0}} & 54.8 & {\color[HTML]{38761D} \textbf{64.5}} & {\color[HTML]{38761D} \textbf{94.5}} & {\color[HTML]{38761D} \textbf{66.6}} & {\color[HTML]{38761D} \textbf{74.2}} & {\color[HTML]{38761D} \textbf{63.7}} & 46.0 \\ 
 & Entropy 25& {\color[HTML]{38761D} \textbf{93.1}} & {\color[HTML]{38761D} \textbf{82.6}} & 54.1 & {\color[HTML]{38761D} \textbf{64.7}} & {\color[HTML]{38761D} \textbf{94.5}} & {\color[HTML]{38761D} \textbf{67.1}} & {\color[HTML]{38761D} \textbf{74.4}} & {\color[HTML]{38761D} \textbf{64.2}} & 47.6 \\\hline
\multicolumn{2}{l}{\textbf{Average AUC}}& {\color[HTML]{38761D} \textbf{74.1}} & {\color[HTML]{38761D} \textbf{73.1}} & 52.0 & 59.0 & {\color[HTML]{38761D} \textbf{69.8}} & {\color[HTML]{38761D} \textbf{63.5}} & {\color[HTML]{38761D} \textbf{62.4}} & {\color[HTML]{38761D} \textbf{62.8}} & 55.1 \\ 
\hline\multirow{2}{*}{PPL\_200} & Seen& 1.4$\pm$0.1 & 1.5$\pm$0.1 & 1.7$\pm$0.2 & 2.1$\pm$0.8 & 2.8$\pm$0.3 & 3.1$\pm$0.4 & 3.2$\pm$0.4 & 3.3$\pm$0.6 & 8.1$\pm$2.3 \\ 
 & Unseen& 3.0$\pm$1.6 & 2.0$\pm$0.3 & 1.8$\pm$0.2 & 3.6$\pm$2.4 & 3.8$\pm$0.9 & 4.3$\pm$1.0 & 4.1$\pm$1.1 & 4.6$\pm$1.6 & 9.4$\pm$2.7 \\ 

\hline
\end{tabular}
}
\caption{Average contamination detection AUC for the \textbf{BioMistral} model,  under different domains within
the Medical-NLU dataset. `PPL\_200' represents the average perplexity $\pm$ STD, from the first 200 tokens within every instance.  The color {\color[HTML]{38761D} green} represents AUCs higher than 60.
}
\label{tab:results_BioMistral}
\end{table*}

\newpage
\subsubsection{Metric Distribution in Histogram}\label{sec:appendix_distribution}
In this section, we present the distributions of different metrics both within domain and across domains. 

\begin{figure}[H]
  \includegraphics[width=0.45\textwidth]{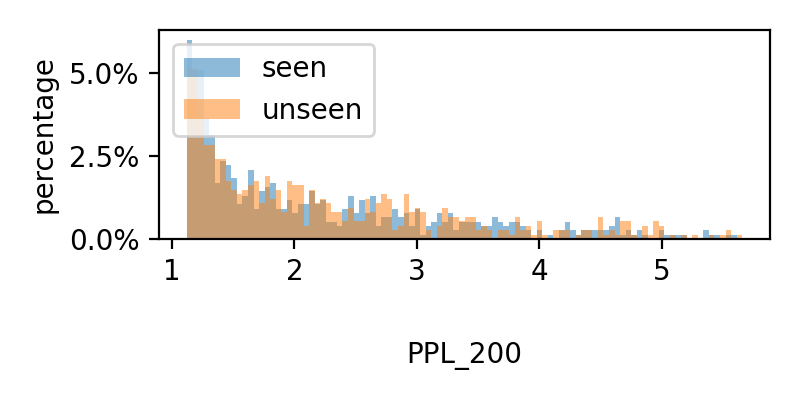}
  \vspace{-5pt}
  \caption{The density plot of \textbf{PPL\_200}  from the Pythia-6.9b model, when both seen and unseen instances are from the \textbf{Github} domain.}
  \label{fig:Github_Github_ppl_200}
\end{figure}

\begin{figure}[H]
  \includegraphics[width=0.45\textwidth]{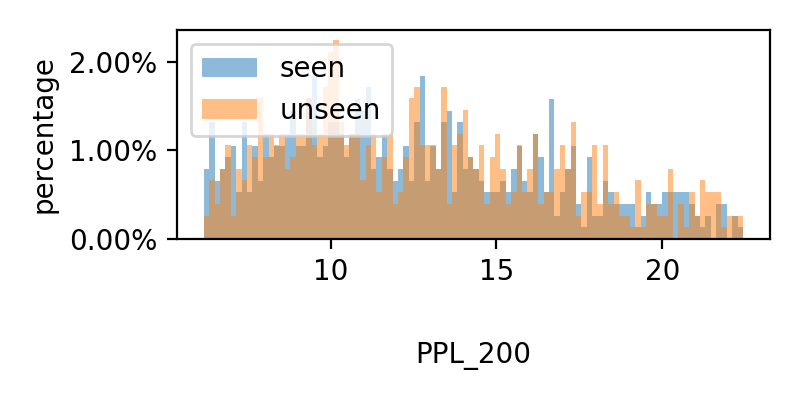}
  \vspace{-5pt}
  \caption{The density plot of \textbf{PPL\_200} from the Pythia-6.9b model, when both seen and unseen instances are from the \textbf{Pile-CC} domain.}
  \label{fig:Pile-CC_Pile-CC_ppl_200}
\end{figure}

\begin{figure}[H]
  \includegraphics[width=0.45\textwidth]{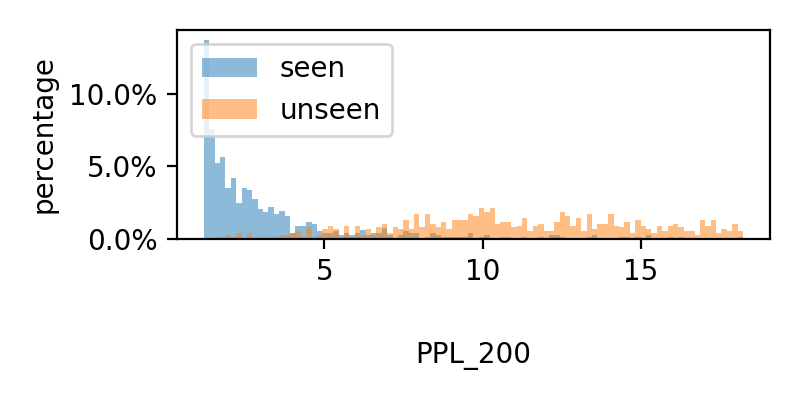}
  \vspace{-5pt}
  \caption{The density plot of \textbf{PPL\_200} from the Pythia-9.6b model, when the seen instances are from the \textbf{Github} domain, and the unseen instances are from the \textbf{Pile-CC} domain.}
  \label{fig:Github_Pile-CC_ppl_200}
\end{figure}

\begin{figure}[H]
  \includegraphics[width=0.45\textwidth]{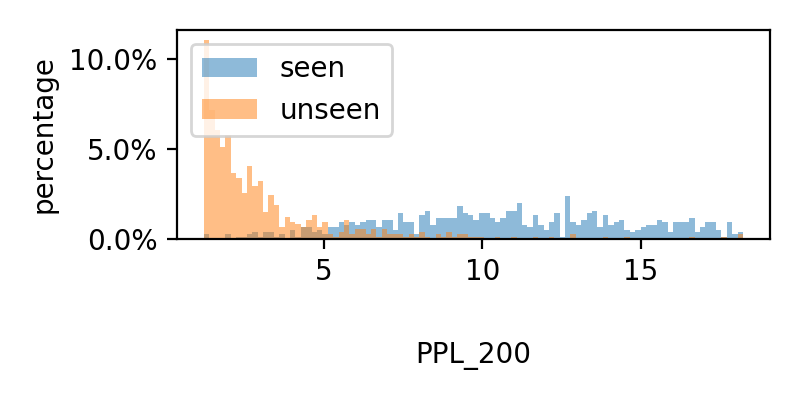}
  \vspace{-5pt}
  \caption{The density plot of \textbf{PPL\_200} from the Pythia-9.6b model, when the seen instances are from the \textbf{Pile-CC} domain, and the unseen instances are from the \textbf{Github} domain.}
  \label{fig:Pile-CC_Github_ppl_200}
\end{figure}

We compare the MIA performance between the GitHub and Pile-CC domains, from the Pythia-6.9b model. As shown in Figure \ref{fig:Github_Github_ppl_200} \& \ref{fig:Pile-CC_Pile-CC_ppl_200}, when the seen and unseen instances are from the same domain, their PPL\_200 distributions are very similar. However, as shown in Figure \ref{fig:Github_Pile-CC_ppl_200} \& \ref{fig:Pile-CC_Github_ppl_200}, when the seen and unseen instances are from different domains, their PPL\_200 distributions are very different. This indicates that the PPL\_200 relates more to domain shifts, instead of the contamination status of individual instances.

\begin{figure}[H]
  \includegraphics[width=0.45\textwidth]{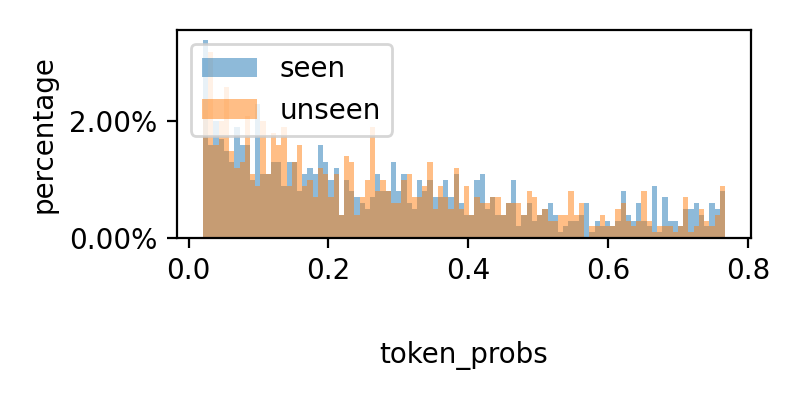}
  \vspace{-5pt}
  \caption{The density plot of \textbf{Min 25\% Prob} when both seen and unseen instances are from the \textbf{Github} domain.}
  \label{fig:Github_Github_min25prob}
\end{figure}

\begin{figure}[H]
  \includegraphics[width=0.45\textwidth]{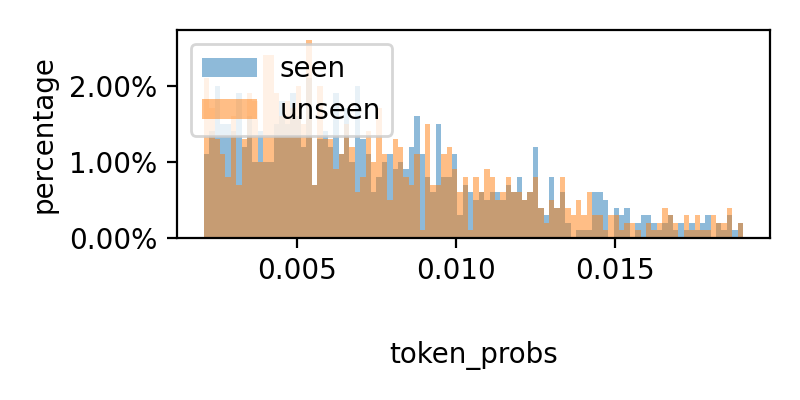}
  \vspace{-5pt}
  \caption{The density plot of \textbf{Min 25\% Prob} when both seen and unseen instances are from the \textbf{Pile-CC} domain.}
  \label{fig:Pile-CC_Pile-CC_min25prob}
\end{figure}

\begin{figure}[H]
  \includegraphics[width=0.45\textwidth]{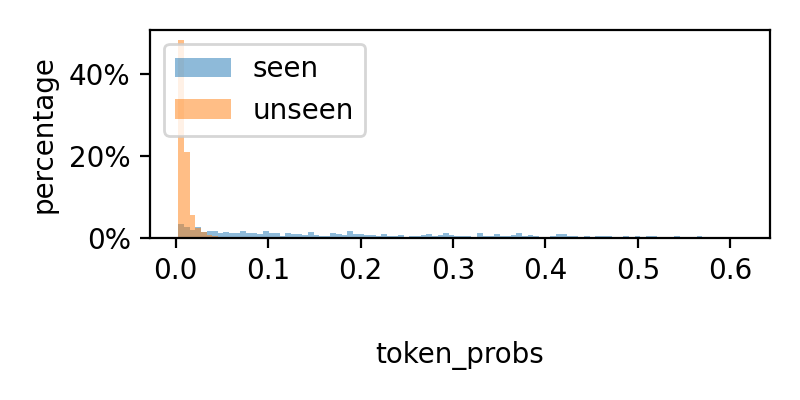}
  \vspace{-5pt}
  \caption{The density plot of \textbf{Min 25\% Prob} from the Pythia-9.6b model, when the seen instances are from the \textbf{Github} domain, and the unseen instances are from the \textbf{Pile-CC} domain.}
  \label{fig:Github_Pile-CC_min25prob}
\end{figure}

\begin{figure}[H]
  \includegraphics[width=0.45\textwidth]{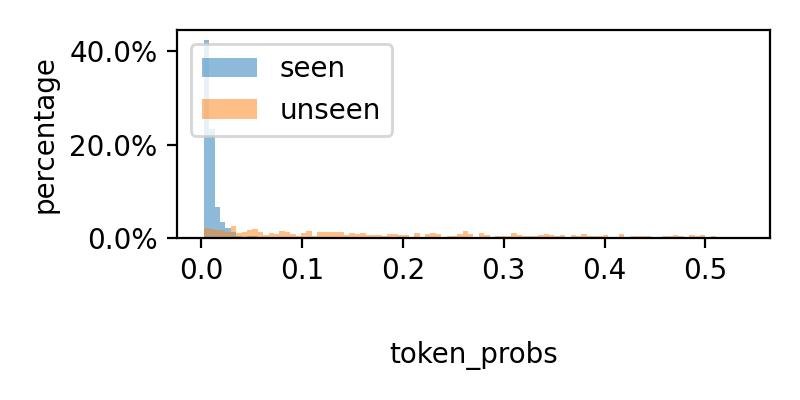}
  \vspace{-5pt}
  \caption{The density plot of \textbf{Min 25\% Prob} from the Pythia-9.6b model, when the seen instances are from the \textbf{Pile-CC} domain, and the unseen instances are from the \textbf{Github} domain.}
  \label{fig:Pile-CC_Github_min25prob}
\end{figure}

\begin{figure}[H]
  \includegraphics[width=0.45\textwidth]{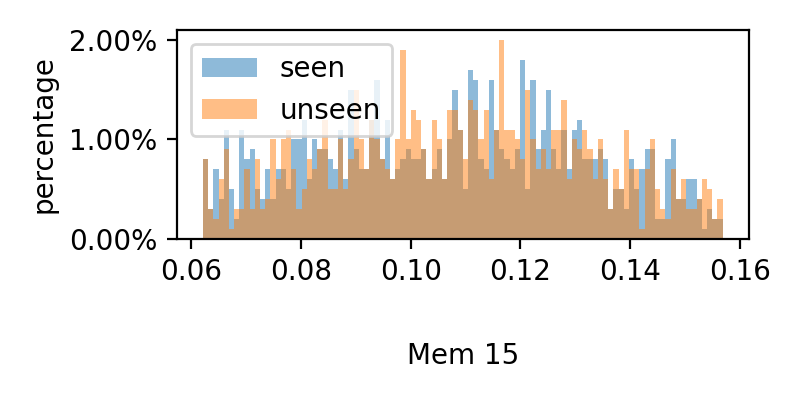}
  \vspace{-5pt}
  \caption{The density plot of \textbf{Mem 25}  from the Pythia-6.9b model,  when both seen and unseen instances are from the \textbf{Github} domain.}
  \label{fig:Github_Github_mem25}
\end{figure}

\begin{figure}[H]
  \includegraphics[width=0.45\textwidth]{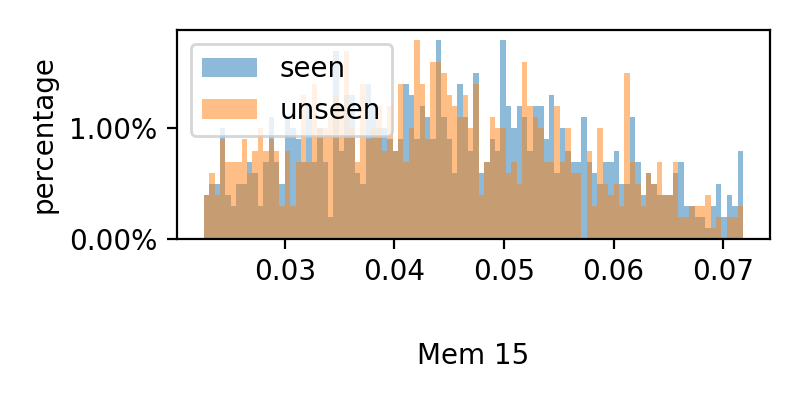}
  \vspace{-5pt}
  \caption{The density plot of \textbf{Mem 25}  from the Pythia-6.9b model,  when both seen and unseen instances are from the \textbf{Pile-CC} domain.}
  \label{fig:Pile-CC_Pile-CC_mem25}
\end{figure}

\begin{figure}[H]
  \includegraphics[width=0.45\textwidth]{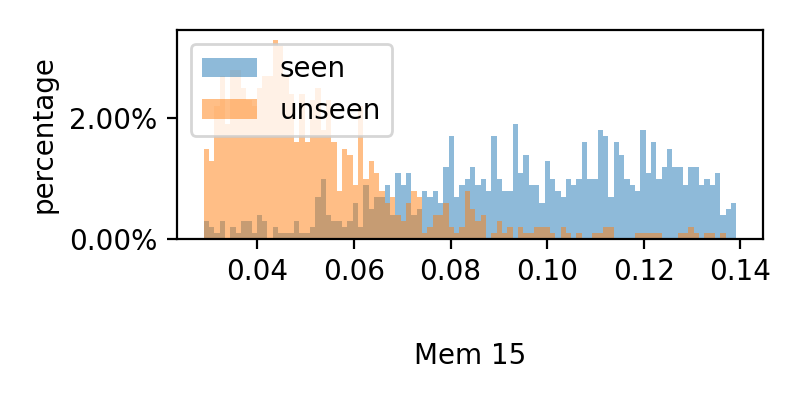}
  \vspace{-5pt}
  \caption{The density plot of \textbf{Mem 25} from the Pythia-9.6b model, when the seen instances are from the \textbf{Github} domain, and the unseen instances are from the \textbf{Pile-CC} domain.}
  \label{fig:Github_Pile-CC_mem25}
\end{figure}

\begin{figure}[H]
  \includegraphics[width=0.45\textwidth]{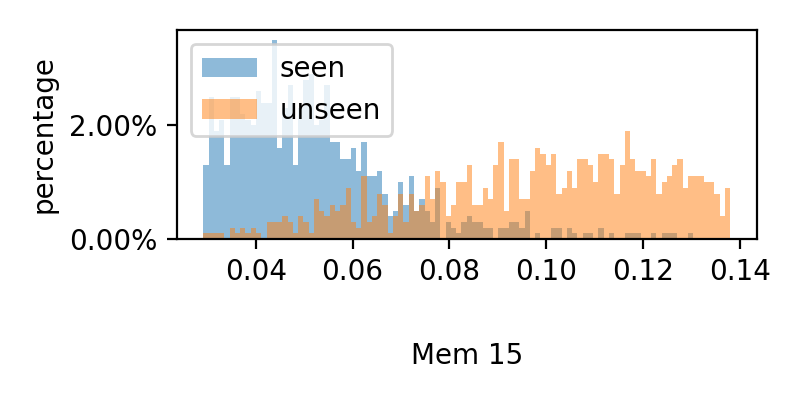}
  \vspace{-5pt}
  \caption{The density plot of \textbf{Mem 25} from the Pythia-9.6b model, when the seen instances are from the \textbf{Pile-CC} domain, and the unseen instances are from the \textbf{Github} domain.}
  \label{fig:Pile-CC_Github_mem25}
\end{figure}

\begin{figure}[H]
  \includegraphics[width=0.45\textwidth]{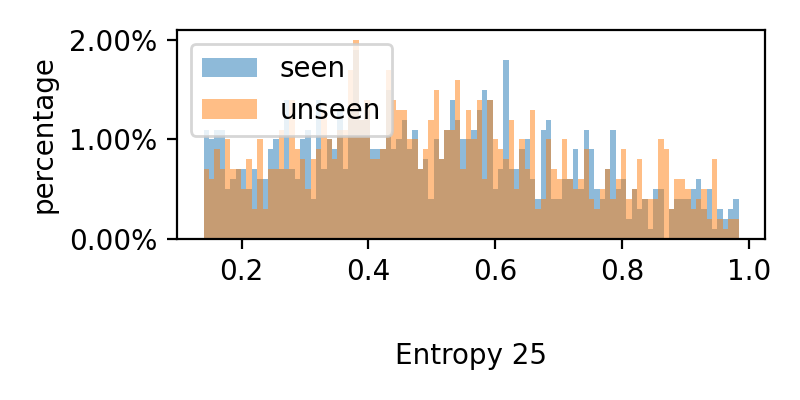}
  \vspace{-5pt}
  \caption{The density plot of \textbf{Entropy 25}  from the Pythia-6.9b model,  when both seen and unseen instances are from the \textbf{Github} domain.}
  \label{fig:Github_Github_entro25}
\end{figure}

\begin{figure}[H]
  \includegraphics[width=0.45\textwidth]{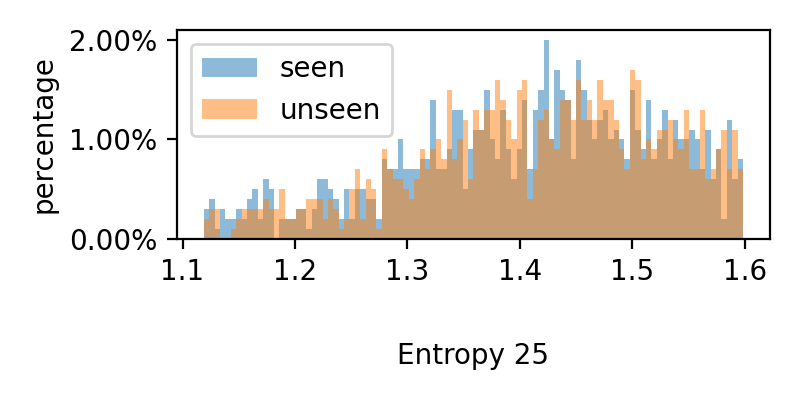}
  \vspace{-5pt}
  \caption{The density plot of \textbf{Entropy 25}  from the Pythia-6.9b model,  when both seen and unseen instances are from the \textbf{Pile-CC} domain.}
  \label{fig:Pile-CC_Pile-CC_entro25}
\end{figure}

\begin{figure}[H]
  \includegraphics[width=0.45\textwidth]{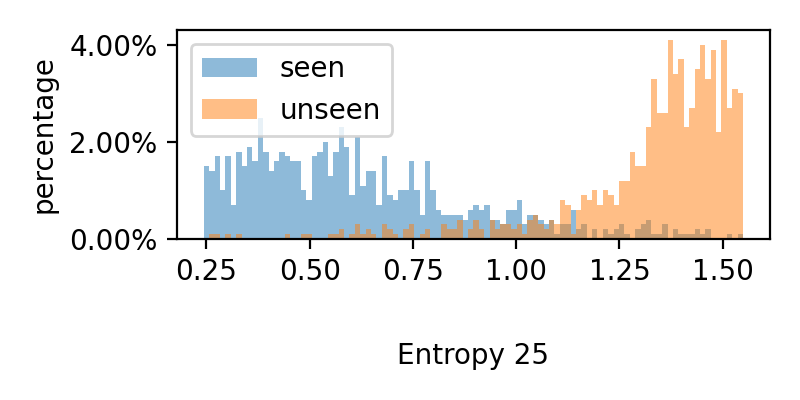}
  \vspace{-5pt}
  \caption{The density plot of \textbf{Entropy 25} from the Pythia-9.6b model, when the seen instances are from the \textbf{Github} domain, and the unseen instances are from the \textbf{Pile-CC} domain.}
  \label{fig:Github_Pile-CC_entro25}
\end{figure}

\begin{figure}[H]
  \includegraphics[width=0.45\textwidth]{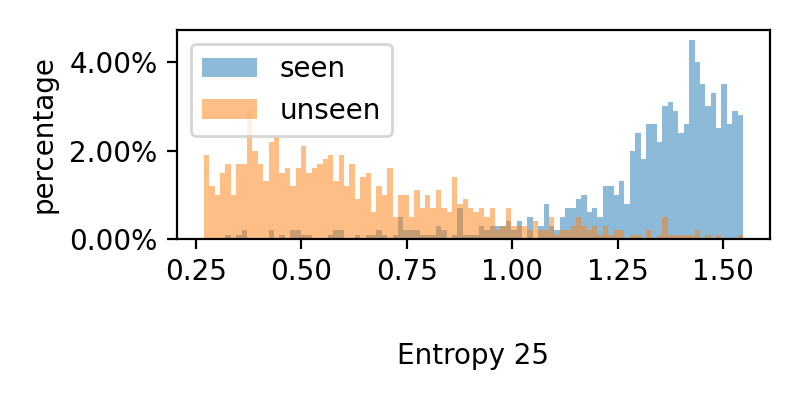}
  \vspace{-5pt}
  \caption{The density plot of \textbf{Entropy 25} from the Pythia-9.6b model, when the seen instances are from the \textbf{Pile-CC} domain, and the unseen instances are from the \textbf{Github} domain.}
  \label{fig:Pile-CC_Github_entro25}
\end{figure}

We observe a similar trend for other metrics: Min 25\% Prob (Figure \ref{fig:Github_Github_min25prob},
\ref{fig:Pile-CC_Pile-CC_min25prob},
\ref{fig:Github_Pile-CC_min25prob},
\ref{fig:Pile-CC_Github_min25prob}),
Mem 25 (Figure \ref{fig:Github_Github_mem25},
\ref{fig:Pile-CC_Pile-CC_mem25},
\ref{fig:Github_Pile-CC_mem25},
\ref{fig:Pile-CC_Github_mem25}),
Entropy 25 (Figure \ref{fig:Github_Github_entro25},
\ref{fig:Pile-CC_Pile-CC_entro25},
\ref{fig:Github_Pile-CC_entro25},
\ref{fig:Pile-CC_Github_entro25}).




\newpage
\subsubsection{Cross-Domain Detection with Different Metrics}\label{sec:appendix_diff_domains}
\begin{figure}[H]
  \includegraphics[width=0.5\textwidth]{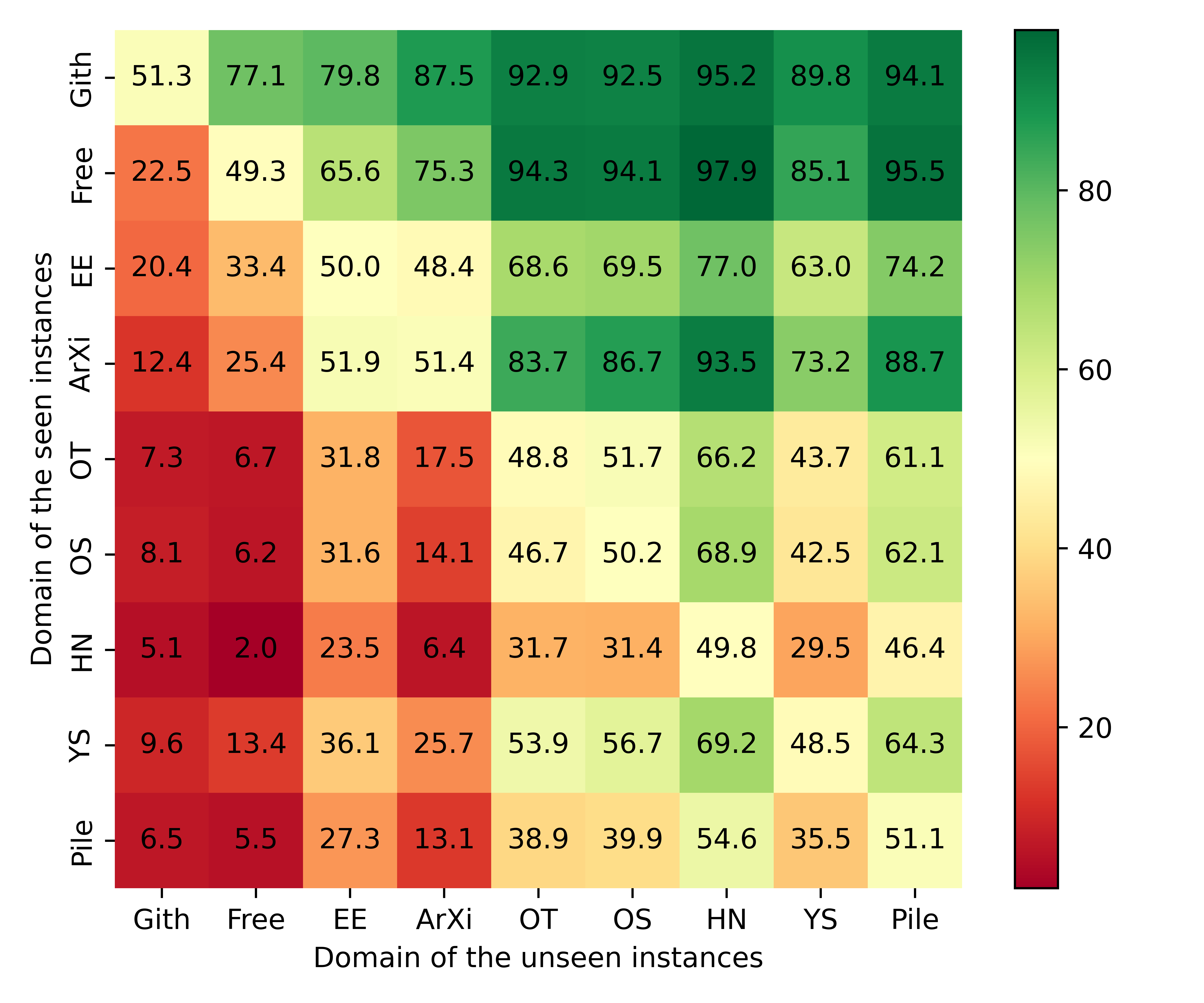}
  \vspace{-5pt}
  \caption{Average contamination detection AUC for the Pythia-6.9b model with the metric, \textbf{Min 25\% token}, when the seen and unseen instances are from different domains. The abbreviations represent the domains in Table \ref{tab:results_pythia-6.9b}. }
  \label{fig:69b_tok25}
\end{figure}

\begin{figure}[H]
  \includegraphics[width=0.5\textwidth]{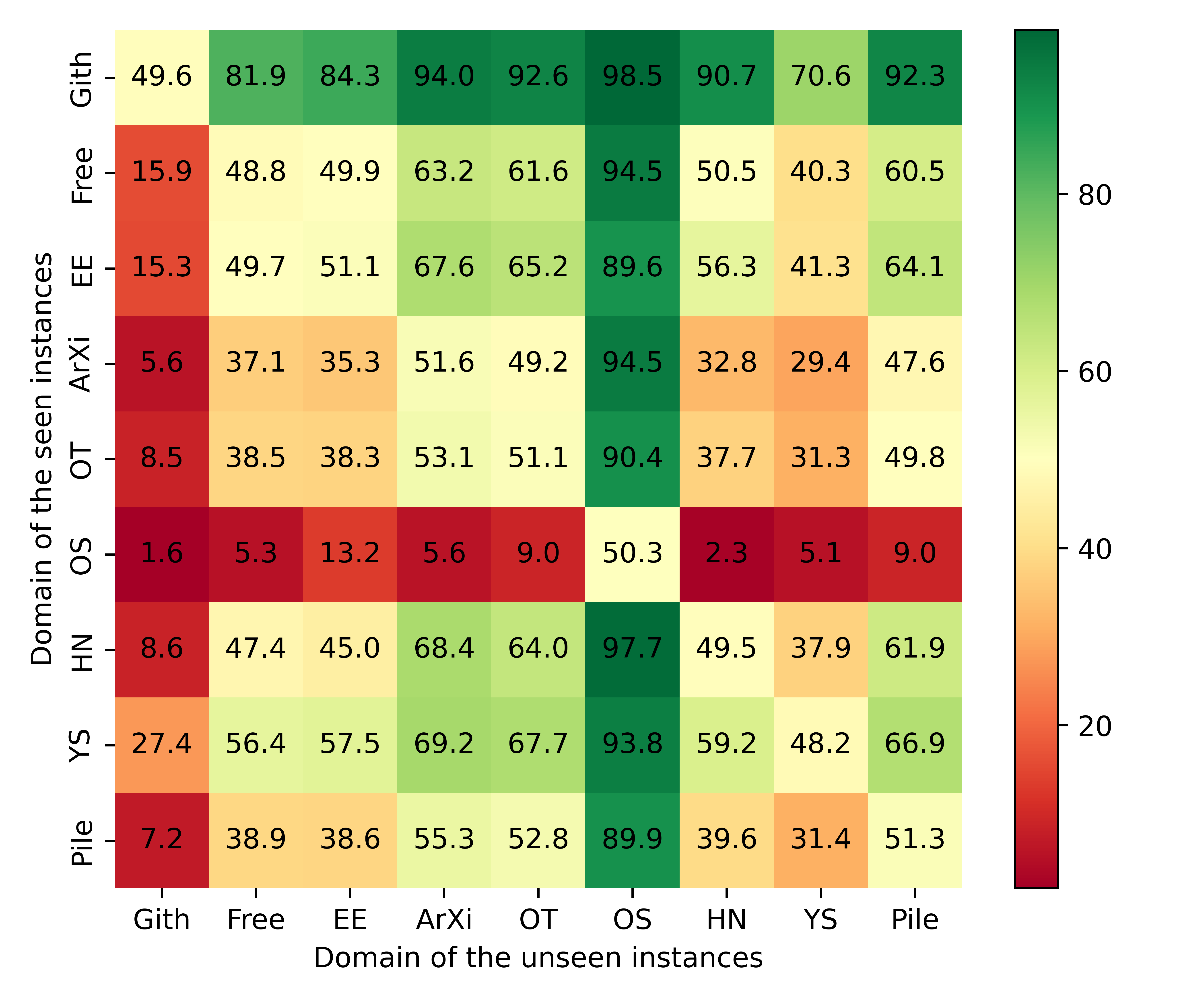}
  \vspace{-5pt}
  \caption{Average contamination detection AUC for the Pythia-6.9b model with the metric, \textbf{Mem 25}, when the seen and unseen instances are from different domains. The abbreviations represent the domains in Table \ref{tab:results_pythia-6.9b}. }
  \label{fig:69b_mem25}
\end{figure}

\begin{figure}[H]
  \includegraphics[width=0.5\textwidth]{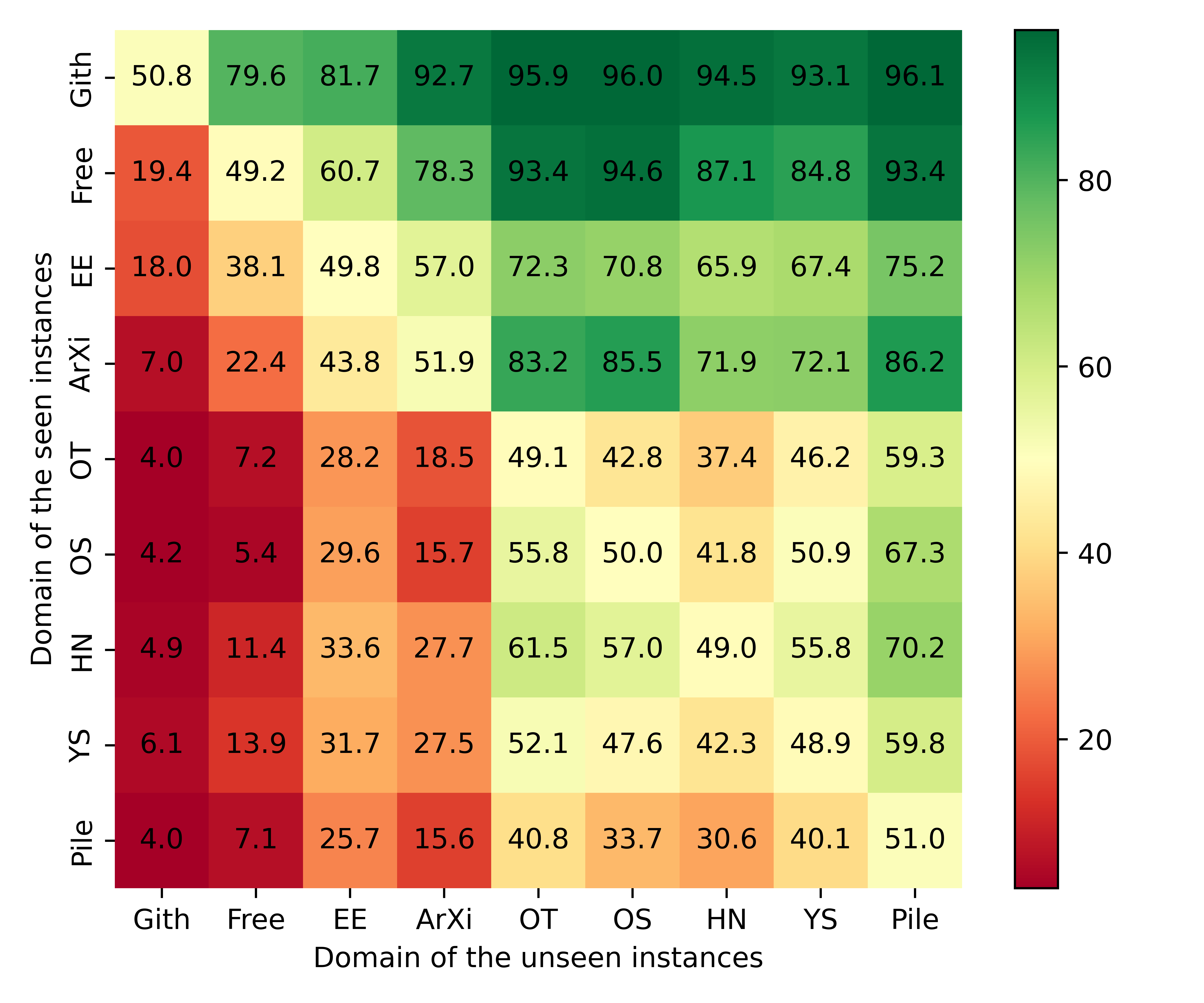}
  \vspace{-5pt}
  \caption{Average contamination detection AUC for the Pythia-6.9b model with the metric, \textbf{Entropy 25}, when the seen and unseen instances are from different domains. The abbreviations represent the domains in Table \ref{tab:results_pythia-6.9b}. }
  \label{fig:69b_ent25}
\end{figure}

In this section, we present AUC results with other metrics,  when seen and unseen instances are from different domains, for the Pythia-6.9b model. The metrics include Min 25\% token (Figure \ref{fig:69b_tok25}), Mem 25 (Figure \ref{fig:69b_mem25}), and Entropy 25 (Figure \ref{fig:69b_ent25}). All metrics exhibit higher AUC values in the top-right corner and lower values in the bottom-left, while diagonal points approach random guessing.

\end{document}